\pdfoutput=1

\documentclass[final,11pt]{article}
\usepackage{acl}

\usepackage{times}
\usepackage{latexsym}
\usepackage{graphicx}
\usepackage{fancyhdr}

\usepackage[T1]{fontenc}

\usepackage[utf8]{inputenc}

\usepackage{microtype}

\usepackage{amsmath}
\usepackage{amssymb}
\usepackage{enumitem}
\usepackage{mdframed}

\usepackage{booktabs}
\usepackage{graphicx}
\usepackage{pifont}
\usepackage{subcaption}
\newcommand{\cmark}{\ding{51}}
\newcommand{\xmark}{\ding{55}}
\usepackage{multirow}
\newcommand\mf[1]{{#1}}
\newcommand\tslnote[1]{{#1}}

\usepackage{booktabs}

\newcommand{\eg}{{\it e.g.}}%

\usepackage[final,inline,nomargin,index]{fixme}  
\FXRegisterAuthor{tl}{atl}{\color{blue}Taesung}

\def\sec{Section}
\def\tab{Table}

\def\fig{Figure}

 \twocolumn 

\title{Matching Pairs: Attributing Fine-Tuned Models to their Pre-Trained Large Language Models}

\author{Myles Foley\thanks{$^*$ Work done during internship at IBM Research.} $^1$, Ambrish Rawat$^2$,   Taesung Lee$^2$,\\ \textbf{Yufang  Hou$^2$,  Gabriele Picco$^2$,   Giulio Zizzo$^2$}\\
  $^1$Imperial College London, 
  $^2$IBM Research \\
  \texttt{m.foley20@imperial.ac.uk} \\
  \texttt{\{ambrish.rawat, yhou\}@ie.ibm.com} \\
  \texttt{\{taesung.lee, gabriele.picco, giulio.zizzo2\}@ibm.com} 
}

\begin{document}
\maketitle

\begin{abstract}

The wide applicability and adaptability of generative large language models (LLMs) has enabled their rapid adoption.
While the pre-trained models can perform many tasks, such models are often fine-tuned to improve their performance on various downstream applications.
However, this leads to issues over violation of model licenses, model theft, and copyright infringement.
Moreover, recent advances show that generative technology is capable of producing harmful content which exacerbates the problems of accountability within model supply chains.
Thus, we need a method to investigate how a model was trained or a piece of text was generated and what their pre-trained base model was.
In this paper we take the first step to address this open problem by tracing back the origin of a given fine-tuned LLM to its corresponding pre-trained base model. We consider different knowledge levels and attribution strategies, and find that we can correctly trace back 8 out of the 10 fine tuned models with our best method.
\end{abstract}


\section{Introduction} 

Recent advancements in pre-trained large language models (LLMs) have enabled the generation of high quality texts
that humans have difficulty identifying as machine generated \cite{jan-2022}. 
While these pre-trained models can perform many tasks in the zero-shot or few-shot settings \cite{gpt-3,schick-schutze-2021-just}, 
such models are often fine-tuned to improve their performance on various downstream applications \cite{peters-etal-2019-tune,pfeiffer-etal-2020-mad}. 
As of May 2023, 
there are more than 
209,000 models 
hosted on Huggingface\footnote{\url{https://huggingface.co/models}} and more than 
12,000 
of them belong to the ``\emph{text generation}'' category. Many generation models are fine-tuned from the open-access pre-trained base models such as XLNet \cite{xlnet}, BART \cite{lewis-etal-2020-bart}, or GPT-J \cite{gpt-j} whose training typically requires significant computational resources.

While the proliferation of text generation models has led to the performance improvement for a wide range of downstream applications such as text summarization and dialogue systems,
it has also been repeatedly shown that these pre-trained or fine-tuned LLMs can facilitate the creation and dissemination of misinformation at scale \cite{https://doi.org/10.48550/arxiv.2112.04359}, and the manipulation of public opinion through false ``majority opinions'' \cite{mann_2021}. 
In response, laws like the EU's Digital Services Act (DSA)\footnote{\url{https://eur-lex.europa.eu/eli/reg/2022/2065/oj}} aim at tackling 
these issues
by enforcing procedural accountability and transparency for responsible use of AI-based technologies. 
These growing demands for AI forensics require the development of methods for establishing model ownership, protecting intellectual property, and analyzing the accountability of any violations.

In this work, we systematically investigate \emph{LLM attribution}, a novel task recently proposed at the first ``Machine Learning Model Attribution Challenge (MLMAC)''\footnote{\url{https://mlmac.io/}}, which aims to link an arbitrary fine-tuned LLM to its pre-trained base model using information such as generated responses from the models. Through LLM attribution, regulatory bodies can trace instances of intellectual property theft or influence campaigns back to the base model.
However, determining attribution for fine-tuned LLMs can be challenging as base models often have similar architectures and overlapping training data. For instance,  \textsc{ThePile} \cite{pile}, a large data set that consists of 22 smaller, high-quality datasets, with a total size of 825 GB, was included into the training data for both GPT-J \cite{gpt-j} and OPT \cite{zhang_opt_2022}. 

In this paper, we cast LLM attribution as a classification problem under two different conditions and develop machine learning solutions to tackle the task. Through extensive experimentation, we examine the trade-offs of different approaches and analyze the influence of the fine-tuning datasets and training iterations on LLM attribution. To the best of our knowledge, this is the first comprehensive study for LLM attribution. We also make the code and models public available for future research\footnote{\url{https://github.com/IBM/model-attribution-in-machine-learning}}.

\section{Related work}\label{sec:related_work}


While there is no systematic study of the fine-tuned LLM attribution task, there are related streams of work in the literature.

\paragraph{Watermarking}
Watermarking is the most prevalent technique to ease the problem of proving model ownership by the first party \cite{tuan-etal-2021-learning, uchida_embedding_2017, chen_deepmarks_2019, le_merrer_adversarial_2020}. When given a particular input, a watermarked model returns a specific output. This allows for protection of 
proprietary 
models by having a designed input-output to always attribute the model to its original source. However, such techniques require specific training modifications which might not be readily available to a third party auditor, but they have given rise to many attacks that can evade or render such measures ineffective \cite{wang_attacks_2019, quiring_forgotten_2018, hitaj_evasion_2019}.

\paragraph{Attribution to training data}
Generative models re-use aspects of their training data in their output at inference \cite{shokri_membership_2017, hisamoto_membership_2020, somepalli_diffusion_2022}. This has given rise to membership inference attacks, and concerns over the re-use of training data in applications like image generation, where copyright infringement may occur \cite{somepalli_diffusion_2022}.

\paragraph{Attribution of LLM outputs}
Another form of attribution attempts to map LLM outputs to identifiable sources \cite{rashkin_measuring_2022}. 
This is crucial when models are required to output factual information such as in search engines \cite{nakano_webgpt_2022}. 
As such,
\cite{rashkin_measuring_2022, menick_teaching_2022}
provide a benchmarks and evaluation frameworks for attribution of LLM outputs that make use of human annotations. %

\section{LLM Attribution}\label{sec:llm_attribution}

\paragraph{Definitions}

Formally, given a vocabulary $\Sigma$, an  LLM $m$ is a function from $\Sigma^N$ to $\Sigma^{N+1}$ where $N$ is the input length.
With auto-regression, $m$ can generate an output of the arbitrary length.
In this paper, we use the ``generate'' functionality implemented in Huggingface transformers library that leverages heuristic generation rules to reduce repeated words and determine when to stop.
The input text expecting a certain type of output is often called a \emph{prompt} (\eg, ``The capital of Canada is''),
and the \emph{response} completes the prompt (\eg, ``Ottawa'').
Thus, for a prompt $p \in \Sigma^N$, an LLM $m$ can be used to obtain a response $m(p)$.

In this work, we consider two collections of LLMs --- the first one is a set $B$ of pre-trained base LLMs, and the second one is a collection $F$ of fine-tuned LLMs.
We assume that 
every model 
$m_f \in F$ 
was
effectively obtained by fine-tuning 
a 
model $m_b \in B$.

\begin{figure}
    \centering
    \includegraphics[width=\columnwidth]{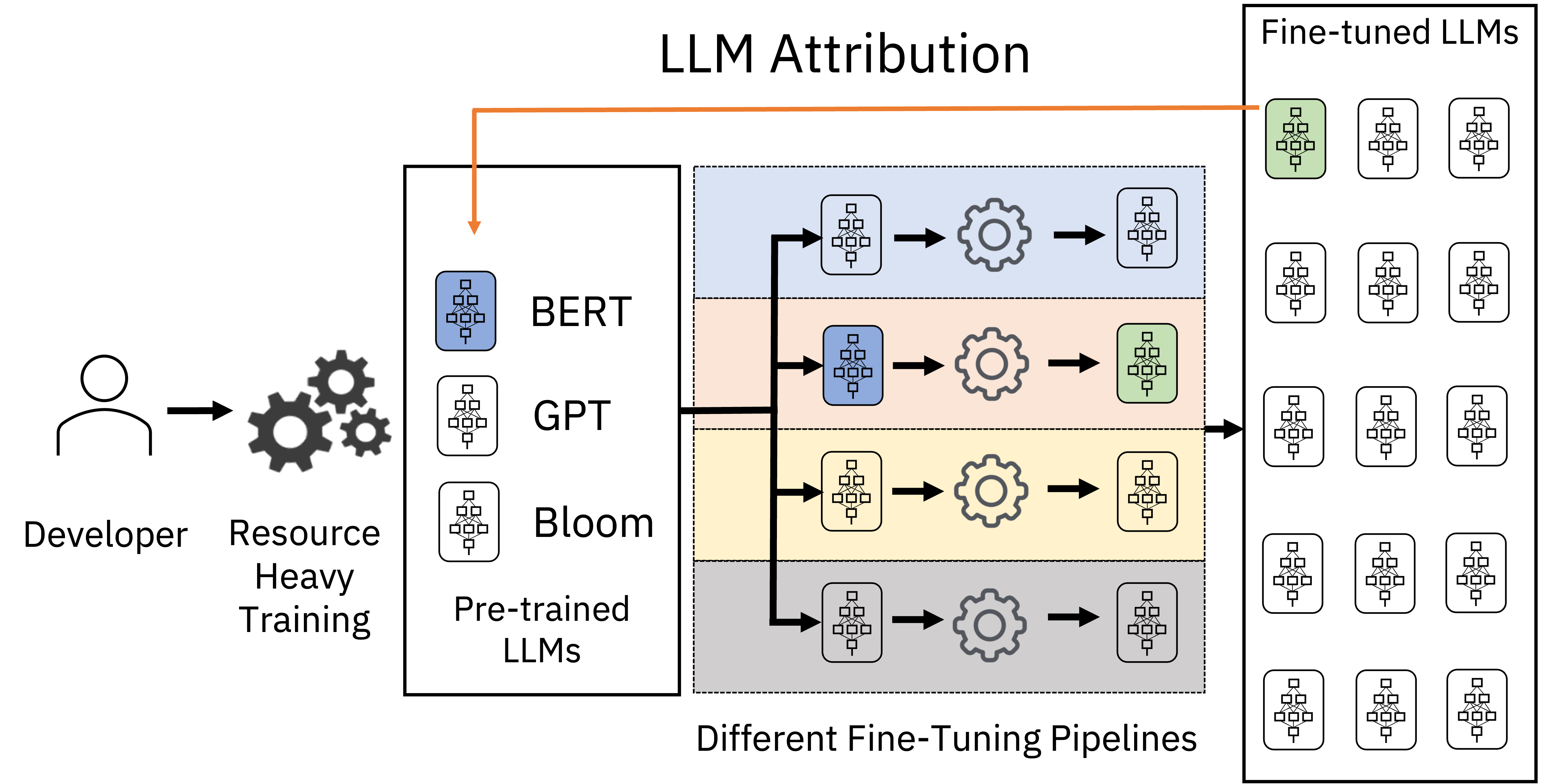}
    \caption{Example of the LLM Attribution Task, illustrating the pipeline of developing $B$, fine-tuing to $F$ and attribution of $m_f$ to $m_b$.}
    \label{fig:attribution}
\end{figure}

\paragraph{Problem Formulation \& Challenges}
The goal of LLM attribution is to design a function $f:F\rightarrow B$ that maps a given fine-tuned model $m_f\in F$ back to its corresponding base model $m_b\in B$ (\fig~\ref{fig:attribution}).
Fine-tuning a given base model $m_b$ involves 
making the choices regarding the fine-tuning dataset and approach. 
This can blur the commonalities between $m_f$ and $m_b$ making attribution an inherently challenging task.
For example, a fine-tuning dataset can be too dissimilar from the pre-training dataset of the base model, or too similar to that of another base model. Similarly, the choice of a fine-tuning algorithm and its hyperparameters, like the number of iterations, can result in catastrophic forgetting \cite{chen-etal-2020-recall}.

The difficulty in designing a mapping $f$ is also tightly linked to the type and degree of allowed access to the models in $B$ and $F$ and the amount of resources available for developing the method.
In general, we assume that the developer of an attribution system can only query the LLMs as 
a black box 
to 
obtain 
the generated responses, and has 
limited 
access to models in $F$.
We speculate this to be true for real-world settings where the producers of pre-trained base models, maintainers of model zoos, or an external auditor are incentivised to develop such attribution systems.
In such scenarios, they may only have limited access to the API of fine-tuned models which will typically be owned by a third-party.
Other constraints may arise from the amount of resources available for developing attribution systems.
For instance, an external auditor may not have the domain expertise or computation resources to benefit from the insights from other fine-tuning pipelines.
Similarly, the developer is assumed to have no knowledge of the fine-tuning datasets and approaches used to obtain the models in $F$, as in these cases attribution may be easily achieved by replicating the setup locally and comparing the obtained models with those in $F$.
In addition to these assumptions,
we consider the following two knowledge levels available with the developer of an attribution system.
\begin{itemize}
    \item \textbf{Universal knowledge} $K_U$: This allows the developer access to universal knowledge about models in $B$. This allows the analysis by a human expert, as well as computing the perplexity of the input. Moreover, the developer can build an additional set of fine-tuned models $A$, or even the capability to train such models. This enables building a supervised attributor using $A$ as a training dataset.
    \item \textbf{Restricted knowledge} $K_R$: We do not have access to $A$, and can only query the models in $B$ as a black box to get the responses.
\end{itemize}

\section{Attribution Methods}\label{sec:methods}

We approach the LLM attribution problem as a classification task.
Essentially, LLM attribution requires identifying the certain robust or latent characteristics of a pre-trained base model within the given fine-tuned model.
The fine-tuned model may retain unique aspects in the pre-training data like events and vocabulary of a specific time period, or the fluency of the base model in a certain domain.

\begin{figure}
    \centering
    \includegraphics[width=\columnwidth]{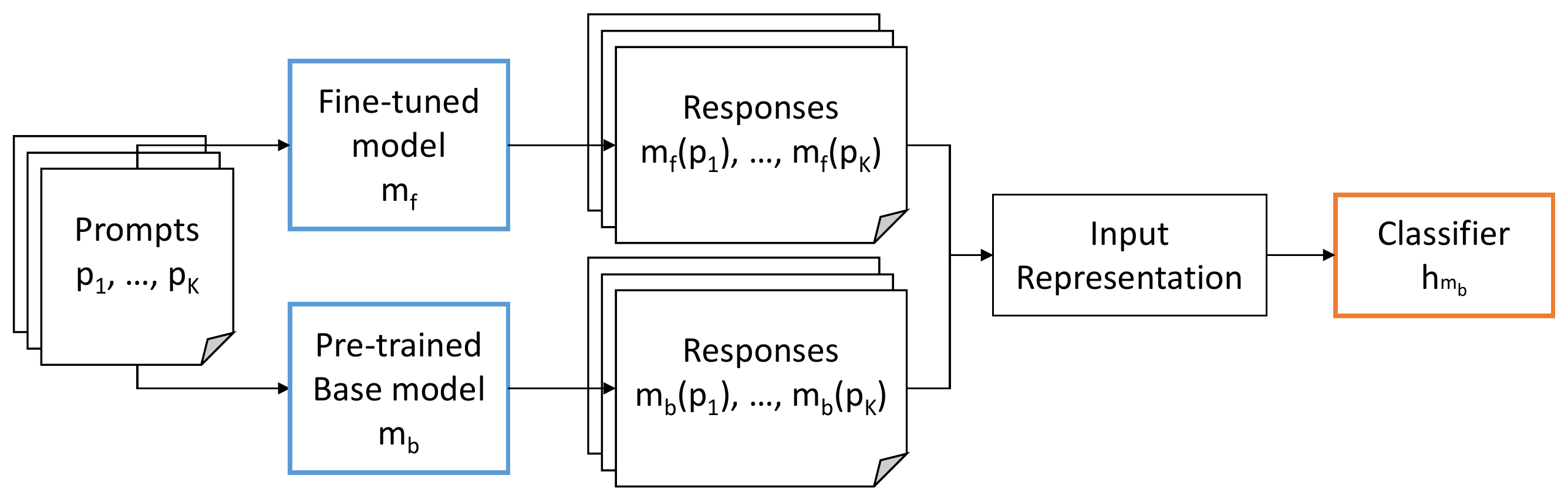}
    \caption{An example configuration of a one-vs-rest classifier $h_{m_b}$ using both base model $m_b$ and fine-tuned model $m_f$.}
    \label{fig:classifier}
\end{figure}

In particular, as shown in \fig~\ref{fig:classifier} we build a classifier $h_{m_b}$ testing a response for each pre-trained base model $m_b$ to decide if a given fine-tuned model $m_f$ retains the characteristics of $m_b$, following the one-vs-rest ($m_b$ or others) scheme.
Then, we aggregate the result to pick the top-1 base model with the majority voting method.
\tslnote{
In other words, we take $m_f$ such that $\sum_{p \in P}{h_{m_b}(m_f(p))}$ is maximized, where $P$ is a set of prompts.
}

The task can be further broken down into two steps for each base model $m_b$ and its classifier $h_{m_b}$ including
(1) characterizing the target base model $m_b$ and representing the input to the classifier (\sec~\ref{sec:methods:input}),
(2) selecting the prompts (\sec~\ref{sec:methods:prompt}),
and (3) designing the classifier (\sec~\ref{sec:methods:classifier}).

\subsection{Model Characterization and Input Representation}

In this step, we characterize an LLM (fine-tuned or base model), and prepare the input to the classifier $h_{m_b}$.
One piece of evidence of attribution lies in exploiting the artefacts of a pre-trained LLM that are expected to persist through the fine-tuning process and are inherited by their fine-tuned counterparts.
For instance, a distinctive feature of RoBERTa \cite{liu_roberta_2019} is the sequence length limit of 512
which is often inherited by its fine-tuned versions.
The task characteristics and associated training data may also help distinguish different LLMs. For example, LLMs trained for specific tasks like chat bots or code generation will have characteristically different output spaces. They may also have unique aspects in their training data like a specific language or markers such as data collected over specific time period.

While 
feature engineering can extract a usable set of features,
it is prone to bias, and less adaptable, and it also requires deep knowledge about $B$.
Thus, we leverage the embeddings of the prompts and responses to learn and exploit such knowledge.

\subsubsection{Input Representation}\label{sec:methods:input}


Our goal is to train a classifier to capture the correlations between an arbitrary response and the base model $m_b$.
For example, with a prompt $p$, this could capture the relationship between a response $m_b(p)$ and $m_b$. Similarly, we can capture the relationship between a response $m_f(p)$ and $m_b$ where $m_f$ is obtained by fine-tuning $m_b$. Assuming that such correlations are preserved in a base model and fine-tuned model pair, we use it to determine the attribution of a fine-tuned LLM.

Given a set of prompts $p_1, \ldots, p_K$, there are multiple ways to prepare them for the classifier.
We can apply the target base model, or fine-tuned model to get the responses, and concatenate the prompt and its response.
Specifically, we list the input representations we consider as follows:
\begin{itemize}
    \item Base model only ($\mathbf{I_B}$): ``$p_i$ $m_b(p_i)$''
    \item Fine-tuned model only ($\mathbf{I_F}$): ``$p_i$ $m_f(p_i)$''
    \item Base model + fine-tuned model ($\mathbf{I_{B+F}}$): ``$p_i$ $m_b(p_i)$ <SEP> $p_i$ $m_f(p_i)$''
    \item Separate embeddings for base model and fine-tuned model.
\end{itemize}
We embed these concatenated sentences using BERT
computed by a \texttt{best}-\texttt{base}-\texttt{multilingual}-\texttt{cased} model\footnote{https://huggingface.co/bert-base-multilingual-cased} except for the last approach 
that embeds 
the components separately for margin-based classifier TripletNet 
described 
in \sec~\ref{sec:methods:classifier}.
Note that all reference to a fine-tuned model $m_f$ during training are actually sampled from another set $A$ of fine-tuned models under $K_U$ assumption as we assume only sparse access to $m_f$.
Also, $I_B$ takes the responses from $m_f$ during prediction to test if the responses share the same characteristics that this classifier learned about $m_b$.


\subsubsection{Prompt Selection} \label{sec:methods:prompt}
While many corpora to pre-train LLMs provide prompts, they might not be all useful to predict the base model.
\mf{Thus, we aim to test and select prompts with more distinctive outcome. Our prompt selection strategy is driven to help best characterise base models. We first collect the list of datasets used in training each base model, identifying unique aspects of datasets that can help identify a base model. Intuitively, one might expect such unique prompts or `edge cases' to bring out the distinctive aspects in the subsequent fine-tuned models. Specifically, we first identify the unique categories of prompts (e.g. different languages) present in different datasets and sample from this set.}\footnote{\mf{Prompt selection is complex and the numerous dimensions of fairness and robustness of such schemes are useful for further investigation of the LLM attribution problem. However, we believe them to be out of scope of this first systematic study on LLM  attribution.}}

More specifically, we consider three approaches:
a small set (\textbf{P1}) of \emph{edge cases} that are distinct to each corpus,
a naive collection (\textbf{P2}) of prompts, and reinforcement learning to select a subset (\textbf{P3}) from the edge cases.


While the naive collection of the 10,000 prompts from ThePile corpus and manually selecting a set of prompts unique to each training dataset 
is clear, we can also use reinforcement learning to optimize the selection using the classification result.
More specifically, we train an agent for each $h_{m_b}$ that can supply prompts for attribution inference. During the training episodes, the agent is rewarded for prompts whose responses lead to correct attribution.
The reinforcement learning setup for this problem is defines as follows:
\begin{itemize}[noitemsep]
    \item \textbf{State.} A feature space consisting of the classification of the prompt, and an embedding of the prompt response computed by \texttt{best-base-multilingual-cased}.
    \item \textbf{Action.} Selecting one of the prompts from \textbf{P1}.
    \item \textbf{Reward.} Using a sparse reward function we reward (+1) for correct classification and penalise (-10) for incorrect classification.
    \item \textbf{Episode.} 20 possible actions.
\end{itemize}
At the start of each episode we are able to randomly select one of the models that the classifier was trained on, thus the RL agent learns to generalise to a variety of different models.
We implement the RL agent using the Proximal Policy Optimisation (PPO) method \cite{DBLP:journals/corr/SchulmanWDRK17}. 

We can use these collected prompts in a few different ways.
A simplistic approach is using each set \textbf{P1}, \textbf{P2} or \textbf{P3} individually.
Another approach \textbf{P1+P2} trains the classifier with \textbf{P2}, and then fine-tune with \textbf{P1} to leverage both of them (\textbf{P3} is already a subset of \textbf{P2}) and we find this is promising in our experiments.
See Appendix~\ref{app:pile_split} for details of the prompts used from \textsc{ThePile} for this combination approach.


\subsection{Classifier Architecture}\label{sec:methods:classifier}

We consider a one vs rest setup where for each base model $m_b$ we train a binary classifier $h_{m_b}: \Sigma^M\rightarrow \{0,1\}$ which takes as input a response $s\in\Sigma^N$, optionally with additional tokens, and predicts a score that reflects its association to the based model $m_b$. 
Single embeddings prepared in \sec~\ref{sec:methods:input} can be straightforwardly used in a simple classifier. We fine-tune the BERT model used for the embedding to make the binary prediction with cross-entropy loss.
Given the one-vs-rest approach the positive samples for an $h_{m_b}$ are repurposed as negative ones for the rest of the classifiers $h_{m_l}$ for $m_l \in B\setminus\{m_b\}$.
The best average score thus obtained is used to establish the attribution for $m_f$. 

We also consider TripletNet~\cite{wei_few-shot_2021} based classifiers that use a margin-based loss function using the separate embeddings of the base and fine-tuned model responses.
The {TripletNet} is able to make predictions by taking in a single sentence, computing the output embedding, and finding the closest embedding from the training set and using the label of the training sentence as a prediction.
The cosine distance between the anchor input, positive example, and negative example are then computed as the loss.
We adopt the margin parameter 0.4 from the original paper \cite{wei_few-shot_2021}.

\section{Experiments}

\subsection{Experiment Setup}
For training the attribution models $h_{m_b}$ we make use of popular text corpora including: GitHub, The BigScience ROOTS Corpus \cite{laurencon_bigscience_2022}, CC-100 \cite{conneau_unsupervised_2020}, Reddit \cite{hamilton_inductive_2018}, and \textsc{ThePile} \cite{gao_pile_2020}. 

We also use a variety of prompt sizes for attribution (150 to 10,000), and datasets (IMDB Reviews \cite{maas-EtAl:2011:ACL-HLT2011}, GLUE \cite{wang_glue_2018}, Tajik OSCAR \cite{2022arXiv220106642A}, and Amazon Multilingual \cite{keung-etal-2020-multilingual}.

\begin{table}[]\centering{\footnotesize
\begin{tabular}{@{}lp{2cm}p{4.2cm}@{}}
\toprule
$m_{\#}$ & Base Model & Fine-tuning dataset                                                                                                                                   \\ \midrule
0        & \href{https://huggingface.co/mrm8488/bloom-560m-finetuned-common_gen}{bloom-350m}                 & common\_gen   \cite{lin_commongen_2020}                                                                                                                               \\
1        & \href{https://huggingface.co/KoboldAI/OPT-350M-Nerys-v2}{OPT-350M}                   & Pike, CYS, Manga-v1                                                                                                                          \\
2        & \href{https://huggingface.co/LACAI/DialoGPT-large-PFG}{DialoGPT-large}             & Persuasion For Good Dataset \cite{wang-etal-2019-persuasion}                                                                                                                 \\
3        & \href{https://huggingface.co/arminmehrabian/distilgpt2-finetuned-wikitext2-agu}{distilgpt2}                 & wikitext2  \cite{merity_pointer_2016}                                                                                                                                  \\
4        & \href{https://huggingface.co/ethzanalytics/ai-msgbot-gpt2-XL}{GPT2-XL}                    & the Wizard of Wikipedia dataset     \cite{dinan_wizard_2019}                                                                                                         \\
5        & \href{https://huggingface.co/dbmdz/german-gpt2}{gpt2}                       & 
Wikipedia dump, EU Bookshop corpus, Open Subtitles, CommonCrawl, ParaCrawl and News Crawl.
\\
6        & \href{https://huggingface.co/wvangils/GPT-Neo-125m-Beatles-Lyrics-finetuned-newlyrics}{GPT-Neo-125m}               & \href{https://huggingface.co/datasets/cmotions/Beatles_lyrics}{Cmotions - Beatles lyrics}                                                                                                                \\
7        & \href{https://huggingface.co/textattack/xlnet-base-cased-imdb}{xlnet-base-cased}           & IMDB \cite{maas-EtAl:2011:ACL-HLT2011}                                                                                                                                 \\
8        & \href{https://huggingface.co/veddm/paraphrase-multilingual-MiniLM-L12-v2-finetuned-DIT-10_epochs}{multilingual-MiniLM-L12-v2} & 
Unknown
\\
9        & \href{https://huggingface.co/giulio98/codegen-350M-multi-xlcost}{codegen-350M}               &  \citet{zhu_xlcost_2022}                              \\ \bottomrule
\end{tabular}}\caption{Fine-tuned models, their original base models and the datasets they are fine-tuned on.}\label{tab:models_data_sum}
\end{table}

To provide a wide didactic range of models for our approaches we utilise 10 pre-trained LLMs to create $B$ \tslnote{and corresponding fine-tuned models (Table~\ref{tab:models_data_sum})}: Bloom \cite{scao_bloom_2022}, OPT \cite{zhang_opt_2022}, DialoGPT \cite{zhang-etal-2020-dialogpt}, DistilGPT2 \cite{sanh_distilbert_2020}, GPT2 \cite{radford_language_nodate}, GPT2-XL  \cite{radford_language_nodate}, GPT-NEO \cite{gpt-neo}, CodeGen \cite{nijkamp_codegen_2022}, XLNET
, MultiLingual-MiniLM \cite{wang-etal-2021-minilmv2}. 
These models provide different architectures, parameter sizes, and tasks to offer a variety of model behaviors.

We consider a one-to-one mapping from $B$ to $F$ (and $A$), thus $F$ and $A$ contain ten models each. 
We utilise open-source models that are implemented in the Huggingface library to form the sets of $F$ and $A$. We select $A$ and $F$ such that the fine-tuning dataset, and model configuration are known to us, of these we select the most popular by number of downloads.
We provide further details of these in Appendix~\ref{app:finetuned_models}.

We take the top-1 result for each $m_b$ as mentioned in \sec~\ref{sec:methods} and check its correctness.
We use F1 and ROC curves as additional metrics. These are calculated using prompt-level attribution calculated per $m_b$ (as in Figure~\ref{fig:roc_auc}), and we use an average per $h_{m_b}$ (as in Figure~\ref{fig:average_auc}). Each of the attributors $h_{m_b}$ described is ran once to determine the attribution of $m_f$ to $m_b$. Training is conducted using a single NVIDIA A100 GPU.

\subsection{Compared Approaches}


We consider different configurations for BERT classifiers based on the input representations $\mathbf{I_B}$, $\mathbf{I_F}$ or $\mathbf{I_{B+F}}$, and
the prompts used \textbf{P1}, \textbf{P2}, \textbf{P3} or \textbf{P1+P2} described in \sec~\ref{sec:methods:input}.

We also consider the margin classifier TripleNet (\sec~\ref{sec:methods:classifier}), and the following heuristic approaches.
\begin{itemize}
\item \textit{Perplexity}: A measure of how confident a model is at making predictions, this can be leveraged for measuring attribution by computing the perplexity of $m_b$ relative to the response of $m_f$ to prompt $p$. 

\item \textit{Heuristic Decision Tree (HDT)}: Using $K_U$ we can use knowledge of $B$ to create a series of discriminative heuristics to categorise $F$ as used by the winning solution to the first MLMAC\footnote{https://pranjal2041.medium.com/identifying-pretrained-models-from-finetuned-lms-32ceb878898f}.




\item \textit{Exact Match}: Attribute responses $m_f$ to $m_b$ when both models respond the same to a prompt. Using the argmax of these attributions to attribute $m_f$ to $m_b$.




\end{itemize}
For detailed descriptions of the heuristic approaches, please refer to Appendix~\ref{app:heuristics}.

\subsection{Attribution Accuracy}\label{ssec:attribtuion_acc}
Here, we examine the attribution abilities of the compared approaches shown in \tab~\ref{tab:attribtuion_scores}. 
Under $K_U$ conditions the baselines of Perplexity and HDT are only able to correctly attribute 1 and 5 models respectively. 
Perplexity fails to capture the subtly of attribution, as repetitive responses lead to lower perplexity and so incorrect attribution. The HDT particularly fails to account for overlap in pre-training and fine-tuning. For instance, DialoGPT-Large and $m_{f3}$ (fine-tuned version of distilgpt2) respond in similar short sentences that leads to incorrect attribution.
The TripletNet baseline performs poorly, only correctly attributing 3 of the models.
Both BERT based attributors are able to attribute more models correctly in comparison to the baselines.  

Examining the models at $K_R$ shows similar performance. The exact match correctly attributes 5 models and BERT+$I_B$ identifies 6 models. BERT+$I_B$+$P1+P2$ attributor is the most successful by correctly attributing 8 models.
Note that this model is the most expensive to train as we have to query a large number of prompts.
\begin{figure}
    \centering
    \includegraphics[width=\columnwidth]{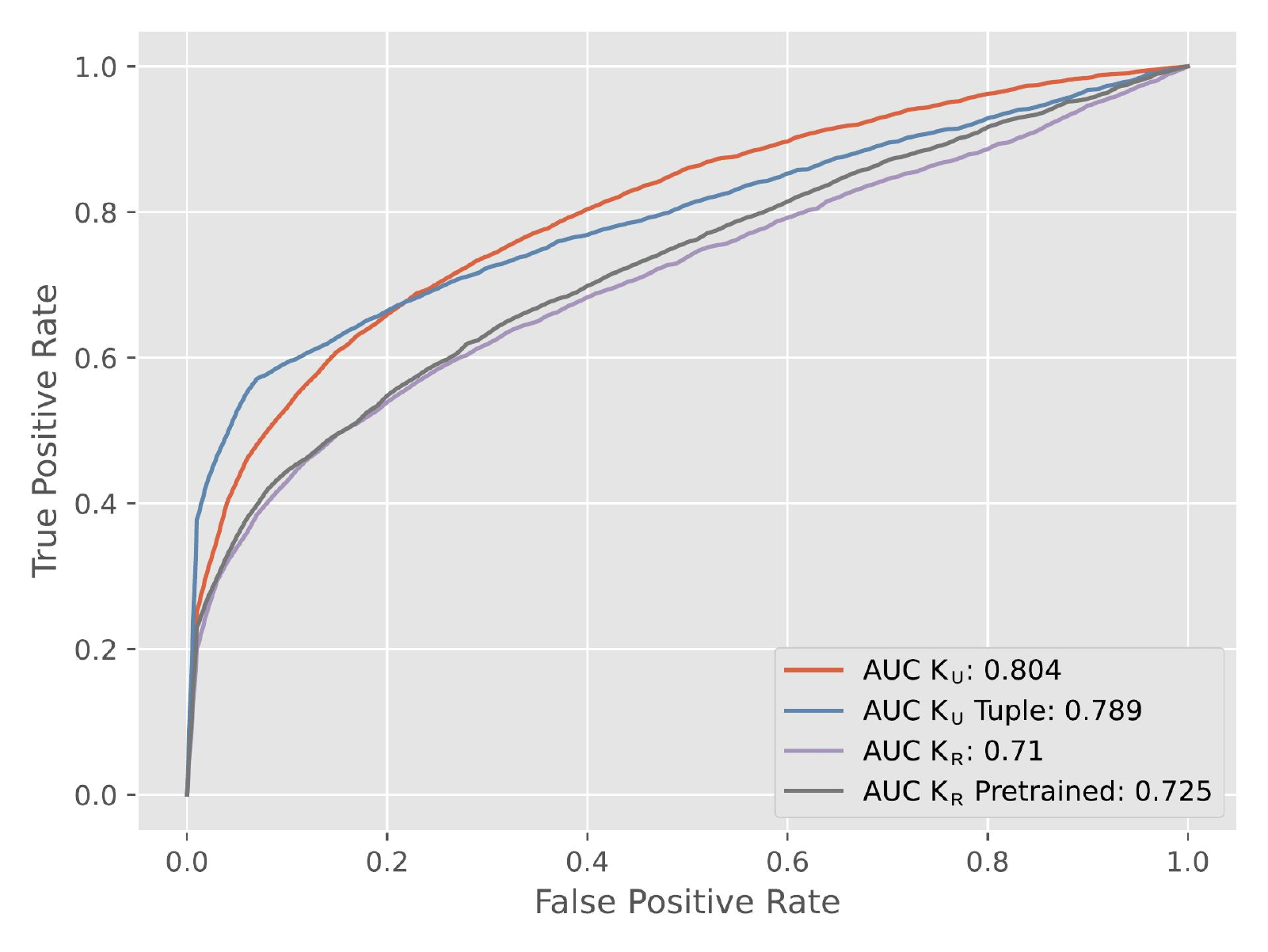}
    \caption{Average ROC plots for each classifier at each knowledge level.}
    \label{fig:average_auc}
\end{figure}

We compare the ROC curves for BERT based attributor defined under each $K$ in Figure~\ref{fig:average_auc}. We provide plots of $h_{m_b}$ in each variant in Appendix~\ref{app:auc}. 
It is interesting to note that the models under $K_R$ have shallower curves than their $K_U$ counterparts, yet these $K_R$ models lead to the same or higher number of correct attributions.
This is likely due to the `noise' that gets added to responses of $A$ from their separate fine-tuning task, $\mathcal{T_A}$. This noise moves the responses of $m_a$ further from $m_f$ (and by extent $m_b$). As such responses from $m_b$ are closer to $m_f$ than $m_a$.
This makes the attributors predict more negative samples correctly under $K_U$ as there is greater disparity in response between $m_a$ and $m_f$, leading to a higher AUC; but also to miss-attribution of $m_f$ at inference. 
Hence, it is unsurprising that the pretrained $K_R$ has the lowest AUC of any model, yet it leads to the highest attribution accuracy in Table~\ref{tab:attribtuion_scores} as it is trained on responses of $m_b$ which is closer in the latent space to responses of $m_f$ than $m_a$.

\begin{mdframed}[backgroundcolor=blue!20] 
\textit{\textbf{Lesson Learned:} Even under reduced knowledge level, pre-training was found to be the factor that contributed to the highest attribution performance.}
\end{mdframed}


\begin{table}[]\centering{\scriptsize
\scalebox{0.9}[1]{
\begin{tabular}{p{2.15cm}p{0.2cm}p{0.05cm}p{0.05cm}p{0.05cm}p{0.05cm}p{0.05cm}p{0.05cm}p{0.05cm}p{0.05cm}p{0.05cm}p{0.05cm}p{0.05cm}p{0.05cm}}
 \toprule
\multicolumn{1}{c}{\multirow{2}{*}{\begin{tabular}[c]{@{}c@{}}Attribution \\ Method\end{tabular}}} & \multirow{2}{*}{$K$} & \multicolumn{10}{c}{$m_{\#}$} & \multirow{2}{*}{TP} \\
\multicolumn{1}{c}{} &  & 0 & 1 & 2 & 3 & 4 & 5 & 6 & 7 & 8 & 9 &  \\ \midrule
HDT & $K_U$ & \cmark & \cmark & \xmark & \xmark & \cmark & \xmark & \xmark & \cmark & \xmark & \cmark & 5 \\
Perplexity & $K_U$ & \xmark & \xmark & \xmark & \xmark & \xmark & \xmark & \xmark & \xmark & \xmark & \xmark & 1 \\
TripletNet + $P1$& $K_U$ & \xmark & \xmark & \cmark & \cmark & \xmark & \xmark & \xmark & \xmark & \xmark & \cmark & 3 \\
BERT + $I_F$ + $P1$ & $K_U$ & \xmark & \cmark & \cmark & \cmark & \xmark & \xmark & \xmark & \cmark & \cmark & \cmark & 6 \\
BERT + $I_{B+F}$ + $P1$ & $K_U$ & \xmark & \cmark & \cmark & \xmark & \xmark & \xmark & \cmark & \cmark & \cmark & \cmark & 6 \\ \midrule
Exact matching & $K_R$ & \cmark & \cmark & \cmark & \xmark & \xmark & \xmark & \cmark & \xmark & \xmark & \cmark & 5 \\ 
BERT + $I_B$ + $P1$ & $K_R$ & \cmark & - & \cmark & - & \xmark & \xmark & \cmark & \cmark & \cmark & \cmark & 6 \\
BERT + $I_B$ + $P3$ & $K_R$  & \cmark & \xmark & - & \xmark & \xmark  &  \cmark& \cmark &  \cmark & -  & \cmark & 5 \\
BERT + $I_B$+$P1$+$P2$ & $K_R$ & \cmark  & \cmark & \cmark & -  & \xmark & \cmark & \cmark &  \cmark& \cmark & \cmark & \textbf{8} \\
\bottomrule
\end{tabular}}}\caption{Model Attributions on $m_{\#}$ from the different methods. Dashes (--) are used when multiple models ($m_f$) are attributed to $m_b$. TP denotes True Positives.}\label{tab:attribtuion_scores}
\end{table}


    \label{fig:my_label}


\subsection{Effects of Prompt usage}\label{ssec:prompt_pile}

The number of prompts available to an attributor for classification can have an influence on the attribution performance: we hypothesize that increasing the number of prompts used results in a clearer signal as to the finetuned to base model attribution.

%



We train BERT attributors under the $K_R$ condition, as the $K_R$ pretrained model performed the strongest. For these experiments we do not use RL prompt selection.

\begin{figure}
    \centering
    \includegraphics[width=\columnwidth]{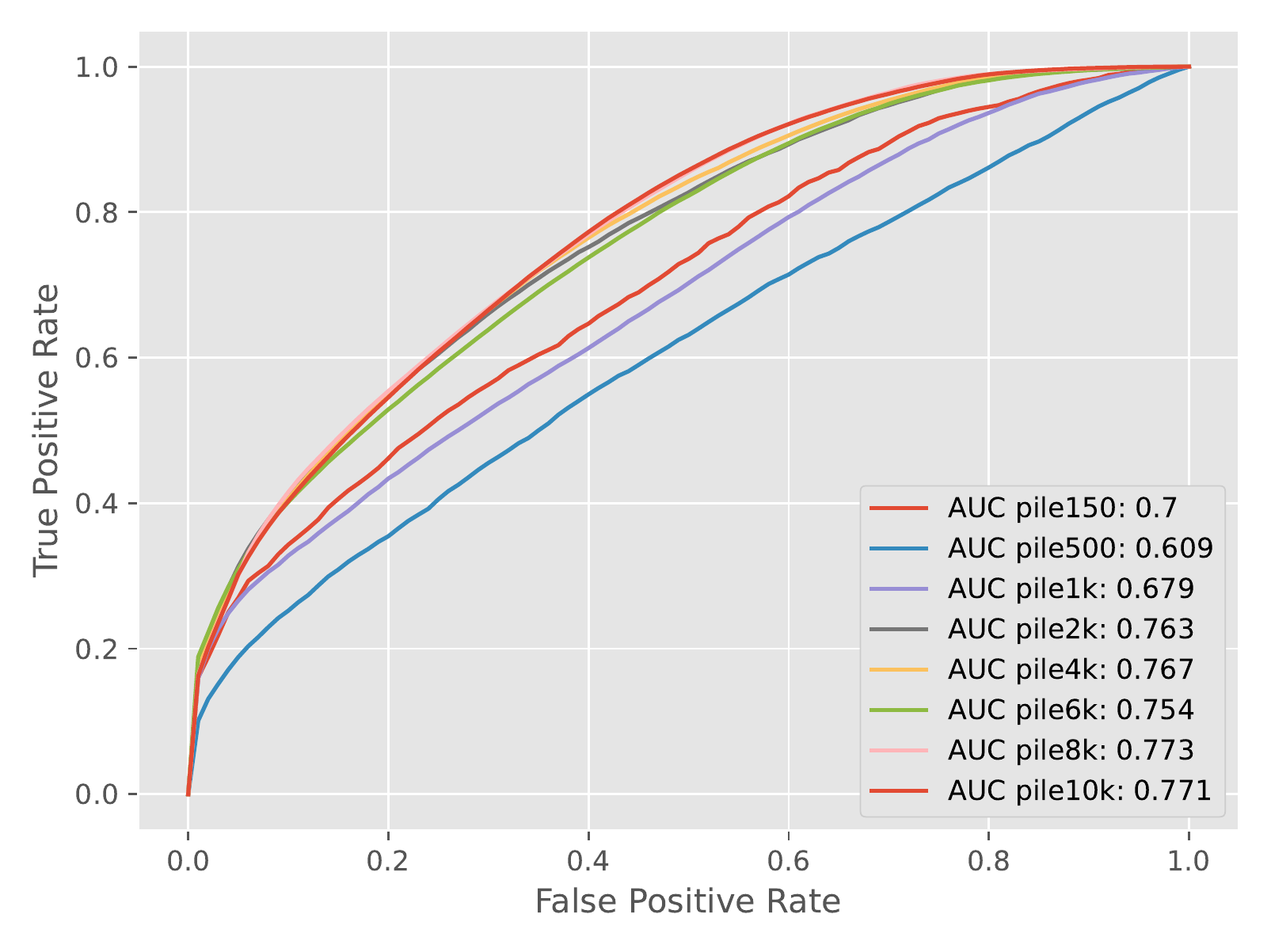}
    \caption{Mean ROC for varying quantities of prompts.}
    \label{fig:average_roc_pile}
\end{figure}

The results of this experiment are shown in Figure~\ref{fig:average_roc_pile}.
By increasing the number of prompts that a classifier is able to use for classification, we see that there is an improvement in the AUC, with diminishing returns from 6,000 prompts onward. 

Increasing the number of prompts improves the AUC, yet does not lead to direct improvement in terms of the attribution accuracy as shown in Table~\ref{tab:train_attribtuion_scores}. In fact, increasing the number of prompts used for classification leads to a highly variable performance. None of the models that directly use these prompts (150 - 10K prompts from the pile) are able to improve or even match that of the pretrained model using 150 prompts from Table 1. 

\begin{mdframed}[backgroundcolor=blue!20] 
\textit{\textbf{Lesson Learned:} Increasing the number of prompts for attribution does not lead to reliable improvements in the number of models correctly attributed.}
\end{mdframed}


\begin{table}[]\centering{\scriptsize
\scalebox{0.9}[1]{
\begin{tabular}{cp{0.05cm}p{0.05cm}p{0.05cm}p{0.05cm}p{0.05cm}p{0.05cm}p{0.05cm}p{0.05cm}p{0.05cm}p{0.05cm}p{0.05cm}p{0.05cm}p{0.05cm}}
\toprule
\multirow{2}{*}{\begin{tabular}[c]{@{}c@{}}Number \\ of Prompts\end{tabular}} & \multicolumn{10}{c}{$m_{\#}$} & \multirow{2}{*}{TP} \\
 & 0 & 1 & 2 & 3 & 4 & 5 & 6 & 7 & 8 & 9 &  \\ \midrule
150 & \xmark & \cmark & \xmark & - & - & - & \cmark & \xmark & - & \cmark & 3 \\
 500& \xmark & \xmark & \xmark & - & - & - & - & - & \cmark & - & 1 \\
1000 & \xmark & \xmark & - & - & - & - & - & \xmark & \cmark & - & 1 \\
2000 & \xmark & \cmark & \cmark & \cmark & - & - & - & \cmark & \cmark & \cmark & 6 \\
4000 & \xmark & \cmark & \cmark & \cmark & \xmark & \xmark & \xmark & \xmark & \cmark & \cmark & 5 \\
6000 & \xmark & \cmark & \cmark & - & - & - & - & \xmark & \cmark & \cmark & 4 \\
8000 & \xmark & \cmark & \xmark & - & - & - & - & \xmark & \cmark &  - & 2 \\
10000 & -  & \cmark & \xmark & - & - & - & - & \cmark  & \cmark & \cmark  & 4 \\
BERT + $I_B$ + $P1+P2$  & \cmark  & \cmark & \cmark & -  & \xmark & \cmark & \cmark &  \cmark& \cmark & \cmark & 8 \\\bottomrule
\end{tabular}}}\caption{Model Attributions on $F$ using a varying number of prompts from The Pile. }\label{tab:train_attribtuion_scores}
\end{table}





\subsection{Effects of pretraining attributors}\label{ssec:pretain_attr}

We next aim to investigate how the size of the pretraining data effects the performance of the attribution, as while using increasingly large data for direct attribution may not improve performance, Section~\ref{ssec:attribtuion_acc} shows that using it as pretraining data does improve attribution. 


To this end each model discussed in Section~\ref{ssec:prompt_pile} is finetuned under $K_R$, varying the size of pretraining data from 150 prompt responses to 10,000. 

\begin{figure*}
\begin{subfigure}{.45\textwidth}
    \centering
    \includegraphics[width=\columnwidth]{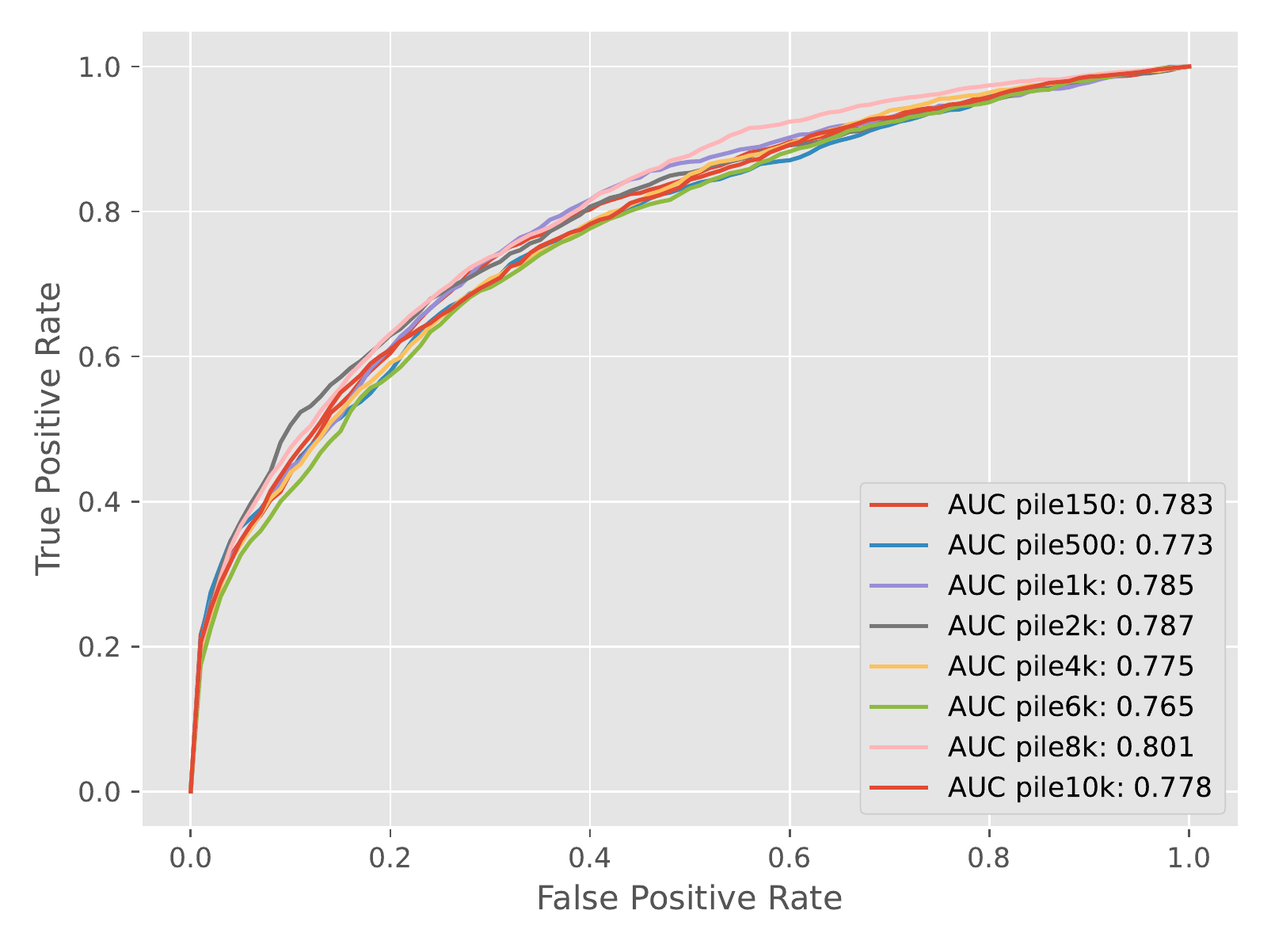}
    \caption{ROC and AUC of attributors with pretrained models using different pretraining data sizes.}
    \label{fig:ft_roc}
\end{subfigure}%
\hfill
\begin{subfigure}{.45\textwidth}
    \centering
    \includegraphics[width=\columnwidth]{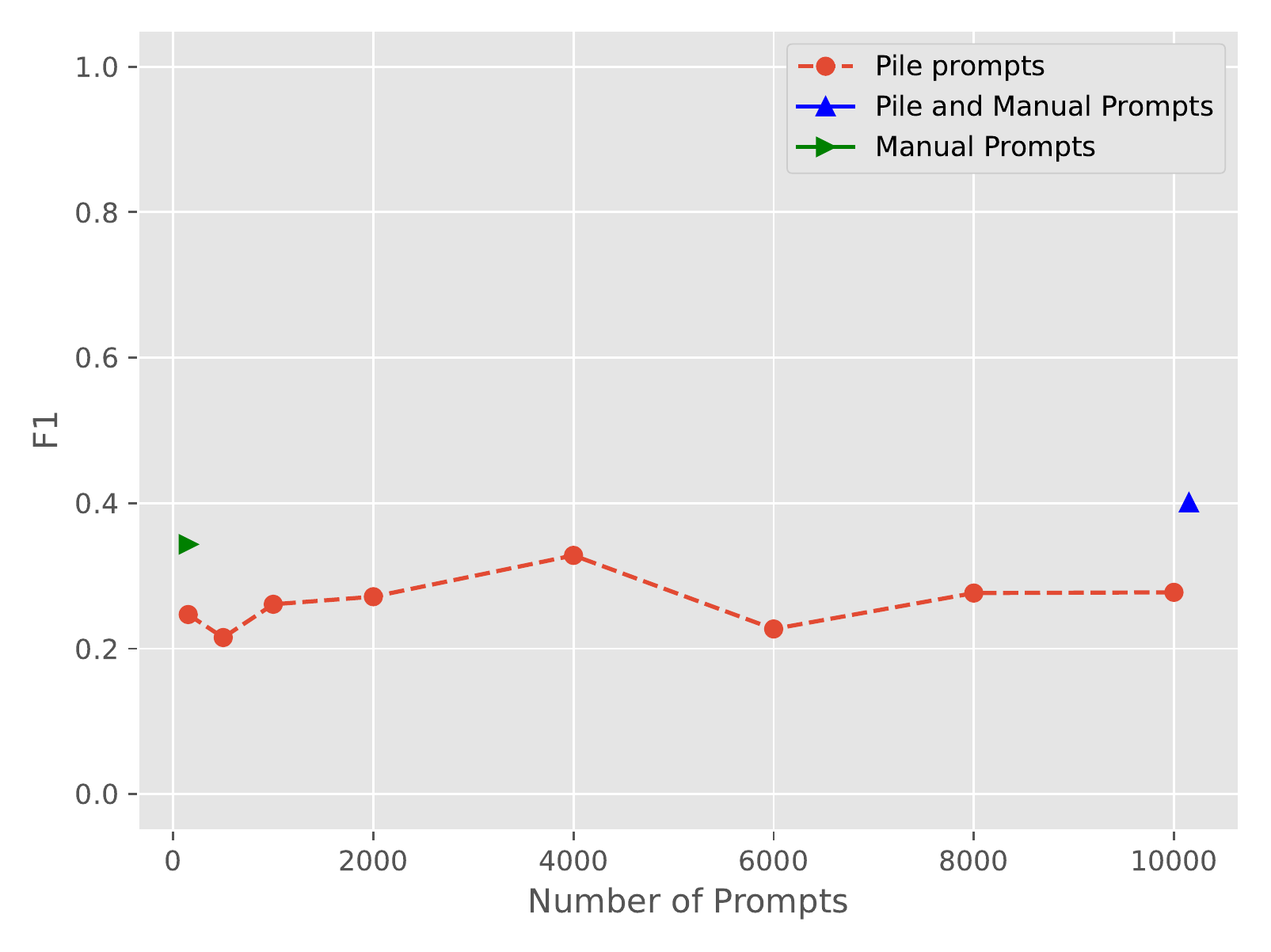}
    \caption{F1 score of attributors with pretrained models using different pretraining data sizes. Manual=150 selected.}
    \label{fig:prec_rec_ft}
\end{subfigure}
\caption{}\label{fig:ft_recall_prec_roc}
\end{figure*}

We report the results of the experiment in Figure~\ref{fig:ft_recall_prec_roc}. In Figure~\ref{fig:ft_roc} we see that the finetuned models are able to improve over the equivalent models in Figure~\ref{fig:average_roc_pile}. Yet they do not improve on the AUC of models trained under $K_U$ conditions.


We see from Figure~\ref{fig:prec_rec_ft} that increasing the number of prompts minimally improves the precision and recall of attribution, with little correlation between number of prompts, even of a varied set like \textsc{ThePile}. Whilst these pretrained-finetuned attributors are able to improve on the precision of the attributor using manual selected prompts, however they are unable to improve on the recall.

What is most important for this task, however, is the ability of attribution, hence we also determine the model attributions for each model in Table~\ref{tab:pretrain_attribtuion_scores}.
The models that have been pretrained on a larger number are able to outperform the $K_R$ model of Section~\ref{ssec:attribtuion_acc} attributing 8 models correctly in the the models pretrained on 4k and 6k prompts. 


\begin{mdframed}[backgroundcolor=blue!20] 
\textit{\textbf{Lesson Learned:} Pretraining attributors is vital to improve the attribution performance. However, this has to diminishing returns in terms of correct attributions and AUC.}
\end{mdframed}


\begin{table}[]\centering{\scriptsize
\scalebox{0.9}[1]
{\begin{tabular}{cp{0.05cm}p{0.05cm}p{0.05cm}p{0.05cm}p{0.05cm}p{0.05cm}p{0.05cm}p{0.05cm}p{0.05cm}p{0.05cm}p{0.05cm}p{0.05cm}p{0.05cm}}
\toprule
\multicolumn{1}{c}{\multirow{2}{*}{\begin{tabular}[c]{@{}c@{}}Number of  \\ Prompts\end{tabular}}} & \multicolumn{10}{c}{$m_{\#}$} & \multirow{2}{*}{TP} \\
\multicolumn{1}{c}{} & 0 & 1 & 2 & 3 & 4 & 5 & 6 & 7 & 8 & 9 &  \\ \midrule
150 & \xmark & - & \cmark & \cmark & \xmark & \xmark & \cmark & \cmark & \xmark & \cmark & 5 \\
500 & \cmark & \cmark & \cmark & \cmark & \xmark & \cmark & \xmark & \cmark & \xmark & \xmark & 6 \\
1000 & \xmark & \cmark & \cmark & \cmark & \xmark & \cmark & \xmark & - & \cmark & \cmark & 6 \\
2000 & \cmark & \cmark & \cmark & \xmark & \xmark & \xmark & \cmark & \cmark & \cmark & \cmark & 7 \\
4000 & \cmark & \cmark & \cmark & - & \cmark & \cmark & \cmark & \cmark & \xmark & \cmark & 8 \\
6000 & \cmark & \cmark & \cmark & - & \xmark & \cmark & \cmark & \cmark & \cmark & \cmark & 8 \\
8000 & \cmark & \cmark & \cmark & \xmark & \xmark & \xmark & \cmark & \cmark & \cmark & \cmark & 7 \\
10000 & \cmark  & \cmark & \cmark & -  & \xmark & \cmark & \cmark &  \cmark& \cmark & \cmark & 8 \\
BERT + $I_B$ + $P1+P2$  & \cmark  & \cmark & \cmark & -  & \xmark & \cmark & \cmark &  \cmark& \cmark & \cmark & 8 \\
\bottomrule
\end{tabular}}}\caption{Model Attributions on $F$ from the models pretrained on different portion from P2, and then fine-tuned with P1.  }\label{tab:pretrain_attribtuion_scores}
\end{table}


\subsection{Effects of Finetuning on Attribution}\label{ssec:effect_finetuning}

The type and duration of the finetuning conducted on a base model $B$ can effect attribution performance. 
To investigate this we use of two base models: distilgpt2 and Multilingual-MiniLM and \mf{finetune them using three datasets:} 
 IMDB \cite{maas-EtAl:2011:ACL-HLT2011}, GLUE \cite{wang_glue_2018}, Amazon reviews Multilingual \cite{keung-etal-2020-multilingual}, and the Tajik language subset of OSCAR \cite{2022arXiv220106642A}.

\mf{Using such datasets more closely models the realistic attack scenario where common pre-training prompt sets are used in an attempt to determine attribution, and fine-tuning datasets are often proprietary and/or unique to the application.
Conducting experiments in this scenario in a controlled setting allows us to study the effect of finetuning on attribution in terms of (a) number of epoch and (b) size of dataset.
}
\begin{figure}
    \centering
    \includegraphics[width=\columnwidth]{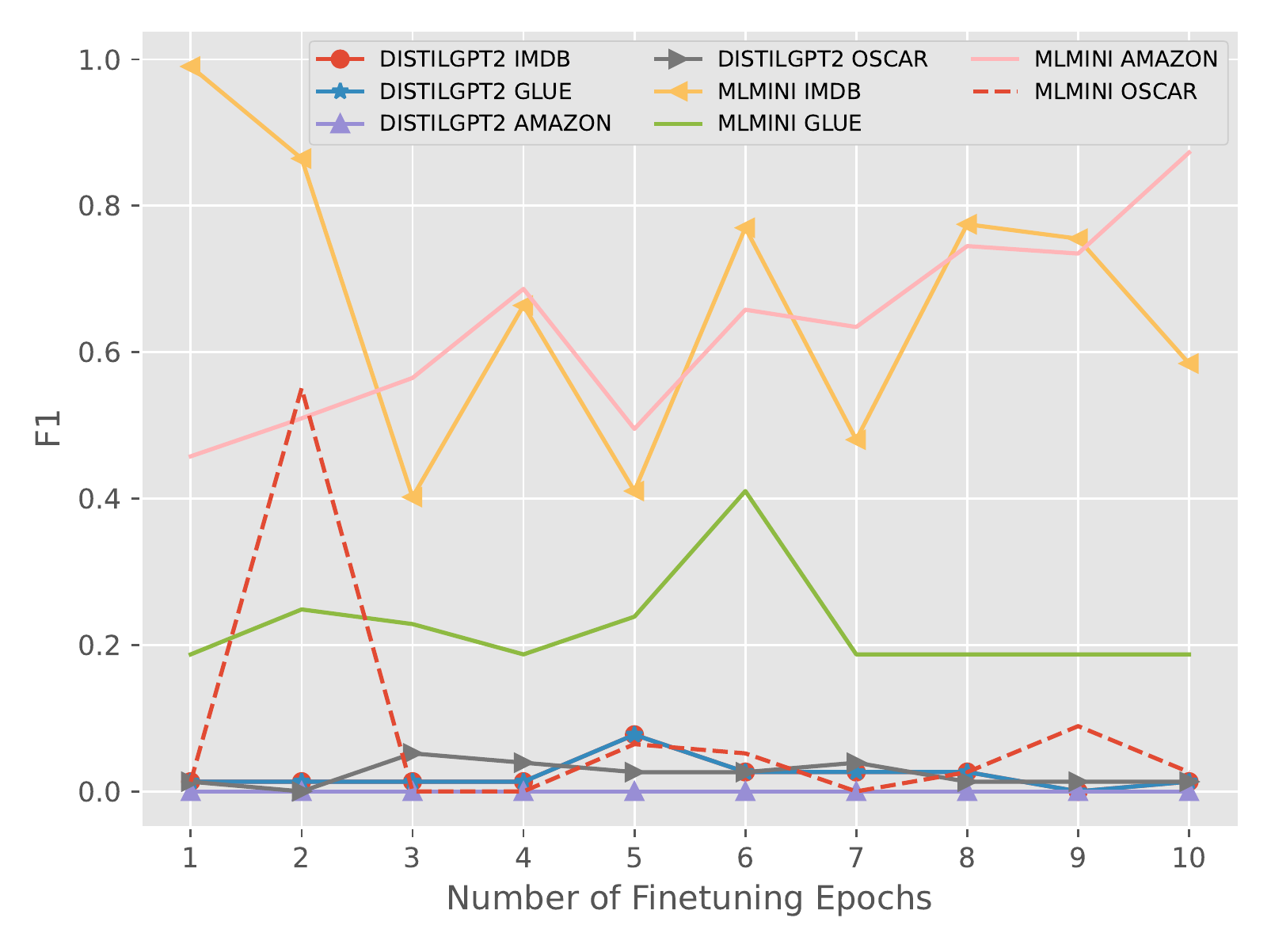}
    \caption{F1 scores of DistilGPT2 and MLMINI attributors.}
    \label{fig:recall_epoch}
\end{figure}

\paragraph{Effect of Finetuning Epochs:} Firstly, we study the effect of the number of finetuning epochs has on attribution. 
Figure~\ref{fig:recall_epoch} shows the F1 score of the MLMini and distilgpt2 attributors when trying to attribute the finetuned base models. 

The MLMini attributor is greatly affected initially by MLMini being finetuned on IMDB, however as with the model finetuned on Amazon reviews there is an increase in attribution performance with increasing finetuning epochs. Conversely, the MLMini model finetuned on GLUE MNLI had minimal change in performance only with anomalous increased F1 score at epoch 6.

However, when trying to attribute MLMINI finetuned with the Tajik subset of OSCAR we see that the F1 score is significantly worse. We speculate that AMAZON and IMDB datasets are similar to the pretraining dataset of MLMini (CC-100) and that the AMAZON reviews, with its 6 languages, are the most similar to this. In fact, the CC-100 is likely to have an overlap in the data distribution of all three of these datasets as all are openly available. As there is no Tajik in CC-100 it is out-of-distribution (OOD) of MLMINI's pretraining dataset, which leads to the poor performance in attribution. 

With the attributor for distilgpt2 there is poor performance in all datasets regardless of the number of epochs. This follows due to the finetuning datasets being OOD relative the the pretraining data of distilgpt2 which used the OpenWebTextCorpus. As OpenWebTextCorpus is mostly in English, finetuning in other languages such as those in the AMAZON dataset, makes attribution harder.

\begin{mdframed}[backgroundcolor=blue!20] 
\textit{\textbf{Lesson Learned:} The attribution performance is dominated by the similarity of the fine-tuning dataset to the pre-training dataset, rather than the amount of fine-tuning conducted.}
\end{mdframed}



\begin{figure}
    \centering
    \includegraphics[width=\columnwidth]{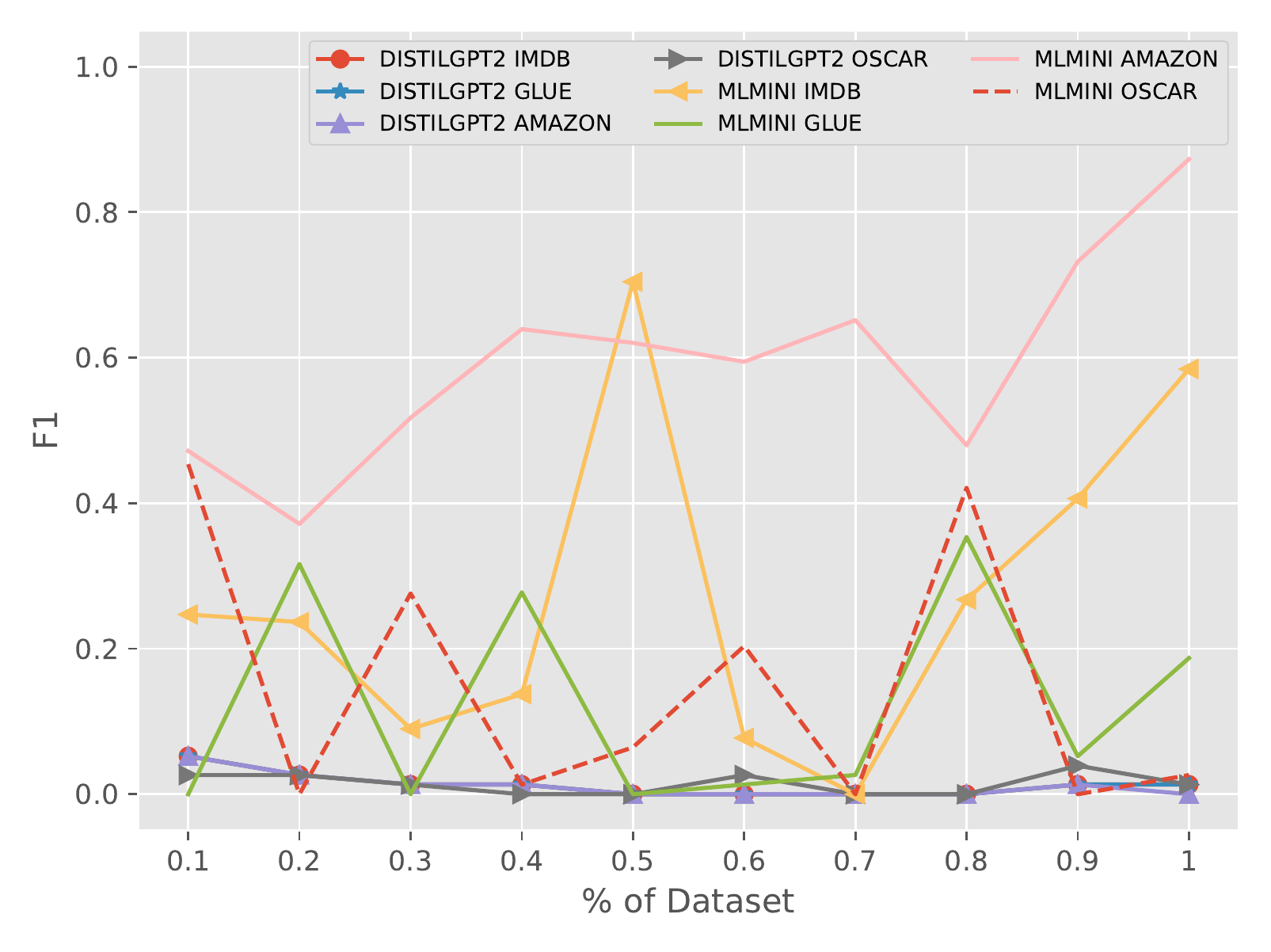}
    \caption{F1 of DistilGPT2 and MLMINI attributors under varying dataset size, relative to original dataset.}
    \label{fig:data_ablation}
\end{figure}

\paragraph{Effects of Dataset Size:}
In addition to the number of finetuning epochs we consider the overall \emph{size} of the finetuning set on attribution. We report the results of using a fixed 10 epochs and varying the finetuning dataset size in Figure~\ref{fig:data_ablation}.  We can see similar effects as in Figure~\ref{fig:recall_epoch}, that the OOD datasets for Distilgpt2 lead to poor F1 scores, 
and consequently, poor attribution results.

For MLMINI we see similar performance on IMDB and AMAZON (two of the in-distibution datasets) with an increased F1 as the dataset size increases. 
When finetuning on OSCAR and GLUE the F1 score shows a minimal correlation with dataset size. 
This again follows from Figure~\ref{fig:recall_epoch}. OSCAR is OOD for MLMINI, which makes attribution significantly harder. Similarly GLUE offers the most varied dataset making attribution harder and giving lower F1. 

\begin{mdframed}[backgroundcolor=blue!20] 
\textit{\textbf{Lesson Learned:} Training on a richer dataset broadly improves results if it is within distribution.}
\end{mdframed}

\paragraph{Effects of Dataset:}
Across Figures \ref{fig:recall_epoch} and \ref{fig:data_ablation} we see the effect of different finetuning datasets on the ability to attribute to base models. 

We can observe the effect of the finetuning datasets on the ability to attribute to base models in Figures \ref{fig:recall_epoch} and \ref{fig:data_ablation}. 
These figures show the distribution of the dataset greatly affects attribution.
Finetuning datasets that are completely out of distribution in relation to the original pre-training dataset severely impact attribution performance. 
This is particularly apparent in MLMINI where finetuning on OSCAR leads to poor attribution performance in Figure \ref{fig:recall_epoch} and \ref{fig:data_ablation}.

Both base models finetuned with GLUE also make attribution harder.
We reason that this is due to the broad range of prompts that are not typical of a finetuning dataset.
This leads the model to produce generic responses to the targeted prompts used for attribution.

\begin{mdframed}[backgroundcolor=blue!20] 
\textit{\textbf{Lesson Learned:} The most significant impact on attribution is the distribution and variety of the finetuing dataset.
}
\end{mdframed}

\section{Conclusion}

In this work we have taken initial steps in the LLM attribution problem. We study LLM attribution in $K_U$ and $K_R$ settings which limit access to $B$ and $F$ to different levels. We argue this prevents trivial solutions in white-box settings, and provides an interesting and realistic study of LLM attribution.

We have considered a variety of different LLMs that are trained on different datasets, and for different purposes. 
We postulate that the 10 different LLMs provide a didactic range of models for LLM attribution. In our experiments, 
we have used pre-existing LLMs that have been fine-tuned by the open-source community to demonstrate the applicability of our methodology. To mitigate the potential for bias this causes, we have tried out best to ensure the fine-tuning task and dataset of such models is known. In addition, we fine-tune a subset of these models in an ablation study, 
which demonstrates the effect that such fine-tuning has on LLM attribution in a controlled environment. 
Our ablation study also studies the effect that OOD fine-tuning datasets have on attribution. This mitigates the effect of only fine-tuning within distribution (of the pre-training data).

Overall, our work contributes to the growing understanding of LLM attribution, laying the foundation for future advancements and developments in this domain.

\newpage
\section*{Limitations}
 
 
 
We have considered a variety of different LLMs in order to study attribution.
However we have only considered a small sample of the different LLM architectures and training strategies.
This has been with a view to using a small but diverse set of LLMs.
Of these 10 base models, we tested our approach to attribution on a controlled set of fine-tuned models. 
While a study that considers a wider variety and larger scale of fine-tuned models would be beneficial to the problem of attribution, the computation resources limited our study.

Furthermore, in our assumptions in this work we consider that there is a one-to-one mapping between $m_f$ and $m_b$. However, this is not necessarily the case. There could be an $m$-to-$n$ mapping and also a model may be present in one set, but not the other.

We believe there is rich space for further research in this area that can address these limitations, and further develop the problem of attribution.

\section*{Ethics Statement}
In the discussion we have highlighted how the techniques for attributing fine-tuned models to their pre-trained large language models can be used as a tool to mitigate issues such as violation of model licenses, model theft, and copyright infringement, but this is only a subset of the issues related to authorship attribution.
The increasing quality and credibility of LLM generated text has recently highlighted ethical issues such as plagiarism
\footnote{\href{https://www.nbcnews.com/tech/chatgpt-can-generate-essay-generate-rcna60362}{New bot ChatGPT will force colleges to get creative to prevent cheating, experts say}}
or the banning of users for submitting AI generated responses to answer questions.
\footnote{\href{https://www.theverge.com/2022/12/5/23493932/chatgpt-ai-generated-answers-temporarily-banned-stack-overflow-llms-dangers}{AI-generated answers temporarily banned on coding Q\&A site Stack Overflow}}
Even within the scientific community discussions are arising related to topics such as the authorship of papers or codes, who owns what is it generated? Many AI conferences have banned the submission of entirely self-generated scientific papers.\footnote{\href{https://www.theverge.com/2023/1/5/23540291/chatgpt-ai-writing-tool-banned-writing-academic-icml-paper}{Top AI conference bans use of ChatGPT and AI language tools to write academic papers}}


These are some examples of controversial situations, but the use of AI-generated content has ethical implications in several domains that depend on the specific context and application. It is therefore crucial, as a first step to tackle these ethical issues, to ensure that any AI-generated contents are clearly labeled as such and are not presented as original work without proper attribution (whether it's a person or a base model).

\section*{Acknowledgements}
This work was supported by European Union's Horizon 2020 research and innovation programme under grant number 951911 – AI4Media.
\clearpage
\newpage
\bibliography{refs,anthology}
\bibliographystyle{acl_natbib}

\appendix

\section{Heuristic Approaches} \label{app:heuristics}

\subsection{Perplexity}

Using the response of $F$ we can calculate the perplexity of $B$ relative to $F$. This can then be used as a measure of how confident $B$ is in predicting $F$, where a lower perplexity would indicate higher confidence and attribution.
In our initial experiments, we found this to be loose approximation of similarity between models in $B$ and $F$. Moreover, this approach assumed stronger access which is typically not available in real-world settings as we discussed in Section \ref{sec:llm_attribution}.

Perplexity is a measure of how well a model is able to predict a sample. It has previously been used in analogous settings for extracting training data from language models \cite{carlini_extracting_2021, mireshghallah_memorization_2022} to determine if a model is confident in its prediction of a sample. It is possible to leverage this for the purpose of attributing $\mathbb{F}$ to $\mathbb{B}$. By collecting responses of $\mathbb{F}$ to prompts we can calculate the perplexity of $\mathbb{B}$ relative to $\mathbb{F}$. Thus we can take the perplexity score as a measure of how confident $\mathbb{B}$ is in predicting the response of $\mathbb{F}$, we would expect lower perplexity to be an indication of higher confidence and therefore higher chances of attribution.

\subsection{Heuristic Decision Tree}

When it comes to generalisation, many LLMs share an equal footing owing to the massive size and intensive training backing their capabilities.
However, when examined closely there are distinctive features that set them apart which can be detected via static or dynamic inspection of the model.
For instance, LLMs with a larger number of parameters tend to take longer for inference. Similarly, length of response varies across LLMs, and some are prone to repetition (such as XLNET \cite{mohamad-zamani-etal-2022-xlnet}). The task characteristics and associated training data may also help distinguish different LLMs. For example, LLMs trained for specific tasks like chat bots or code generation will have characteristically different output spaces. They may also have unique aspects in their training data like a specific language or markers such as data collected over specific time period. Much like watermarking, these can be used to craft prompts that can help reveal these unique artefacts \footnote{\href{https://pranjal2041.medium.com/identifying-pretrained-models-from-finetuned-lms-32ceb878898f}{Winning solution to the first MLMAC}}.

While in principle many of these heuristics can be used for attribution, the practical development of such systems faces a range of challenges. First, these properties may not be preserved across the fine tuning process and therefore provide no meaningful insight for attribution. Second, these heuristics require a high level of expertise and knowledge which may not always be available. An external auditor working with the restricted knowledge of $K_R$ may not be able to develop such solutions. Third, many of the properties of models in $F$ can be easily obfuscated by the exposed API. For example it is fairly easy to normalise response times or post-process the responses to account for repetition. Moreover, an API may be simultaneously backed by multiple different models which would make the attribution even more challenging. Finally, LLMs often have overlapping datasets which can dilute many of the subtleties underlying these heuristics. This limits the applicability and scalability of such approaches for larger collections of $B$ and $F$.


\section{Fine-tuned model Details}\label{app:finetuned_models}
Here we provide details of the fine-tuned LLMs we use in sets $A$ and $F$. Each of the LLMs is an open source implementation hosted on the Huggingface, we provide the link to the fine-tuned model. In Table~\ref{tab:models_data} we show set $F$ as FT models 0-9 inclusive, and set $A$ from 10-19 inclusive. For each model we also provide the dataset used to fine-tune each of the LLMs. 
\begin{table*}[]\centering{\footnotesize
\begin{tabular}{@{}lll@{}}
\toprule
FT model & Base Model               & FT dataset                                                                                                                                   \\ \midrule
0        & \href{https://huggingface.co/mrm8488/bloom-560m-finetuned-common_gen}{bloom-350m}                 & common\_gen   \cite{lin_commongen_2020}                                                                                                                               \\
1        & \href{https://huggingface.co/KoboldAI/OPT-350M-Nerys-v2}{OPT-350M}                   & Pike, CYS, Manga-v1                                                                                                                          \\
2        & \href{https://huggingface.co/LACAI/DialoGPT-large-PFG}{DialoGPT-large}             & Persuasion For Good Dataset \cite{wang-etal-2019-persuasion}                                                                                                                 \\
3        & \href{https://huggingface.co/arminmehrabian/distilgpt2-finetuned-wikitext2-agu}{distilgpt2}                 & wikitext2  \cite{merity_pointer_2016}                                                                                                                                  \\
4        & \href{https://huggingface.co/ethzanalytics/ai-msgbot-gpt2-XL}{GPT2-XL}                    & the Wizard of Wikipedia dataset     \cite{dinan_wizard_2019}                                                                                                         \\
5        & \href{https://huggingface.co/dbmdz/german-gpt2}{gpt2}                       & \begin{tabular}[c]{@{}l@{}}Wikipedia dump, EU Bookshop corpus,\\ Open Subtitles, CommonCrawl, ParaCrawl and News Crawl.\end{tabular}         \\
6        & \href{https://huggingface.co/wvangils/GPT-Neo-125m-Beatles-Lyrics-finetuned-newlyrics}{GPT-Neo-125m}               & \href{https://huggingface.co/datasets/cmotions/Beatles_lyrics}{Cmotions - Beatles lyrics}                                                                                                                \\
7        & \href{https://huggingface.co/textattack/xlnet-base-cased-imdb}{xlnet-base-cased}           & IMDB \cite{maas-EtAl:2011:ACL-HLT2011}                                                                                                                                 \\
8        & \href{https://huggingface.co/veddm/paraphrase-multilingual-MiniLM-L12-v2-finetuned-DIT-10_epochs}{multilingual-MiniLM-L12-v2} & 
Unknown
\\
9        & \href{https://huggingface.co/giulio98/codegen-350M-multi-xlcost}{codegen-350M}               &  \citet{zhu_xlcost_2022}                                                                      \\
10       & \href{https://huggingface.co/wvangils/BLOOM-350m-Beatles-Lyrics-finetuned-newlyrics}{bloom-350m}                 & \href{https://huggingface.co/datasets/cmotions/Beatles_lyrics}{Cmotions - Beatles lyrics}                                                                                                                    \\
11       & \href{https://huggingface.co/Tianyi98/opt-350m-finetuned-cola}{OPT-350M}                   & GLUE \cite{wang_glue_2018}                                                                                                                  \\
12       & \href{https://huggingface.co/mdc1616/DialoGPT-large-sherlock}{DialoGPT-large}             & The complete works of Sir Arthur Conan Doyle                                                                                                                              \\
13       & \href{https://huggingface.co/noelmathewisaac/inspirational-quotes-distilgpt2}{distilgpt2}                 & Quotes-500K    \cite{goel_proposing_2018}                                                                                                                            \\
14       & \href{https://huggingface.co/malteos/gpt2-xl-wechsel-german}{GPT2-XL}                    & OSCAR \cite{2022arXiv220106642A}                                                                                                               \\
15       & \href{https://huggingface.co/lvwerra/gpt2-imdb}{gpt2}                       & IMDB \cite{maas-EtAl:2011:ACL-HLT2011}                                                                                                                                         \\
16       & \href{https://huggingface.co/flax-community/gpt-neo-125M-code-clippy}{GPT-Neo-125m}               & Code Clippy Data dataset   \cite{cooper-2021-code-clippy-data}                                                                                                                  \\
17       & \href{https://huggingface.co/textattack/xlnet-base-cased-rotten-tomatoes}{xlnet-base-cased}           & Rotten Tomatoes \cite{pang-lee-2005-seeing} \\
18       & \href{https://huggingface.co/jegormeister/Multilingual-MiniLM-L12-H384-mmarco-finetuned}{multilingual-MiniLM-L12-v2} & \begin{tabular}[c]{@{}l@{}}\href{https://www.tensorflow.org/datasets/catalog/wikipedia#wikipedia20200301bn}{https://www.tensorflow.org/datasets/catalog/wikipedia}\\ \href{https://www.tensorflow.org/datasets/catalog/wikipedia#wikipedia20200301bn}{\#wikipedia20200301bn}\end{tabular}                        \\
19       & \href{https://huggingface.co/Salesforce/codegen-350M-mono}{codegen-350M}               & BigPython dataset                                                                                                                            \\ \bottomrule
\end{tabular}}\caption{Fine-tuned models, their original base models and the datasets they are fine-tuned on.}\label{tab:models_data}
\end{table*}

\section{AUC Curves}\label{app:auc}
\begin{figure*}
\centering

\begin{subfigure}{.5\textwidth}
  \centering
  \includegraphics[width=0.9\columnwidth]{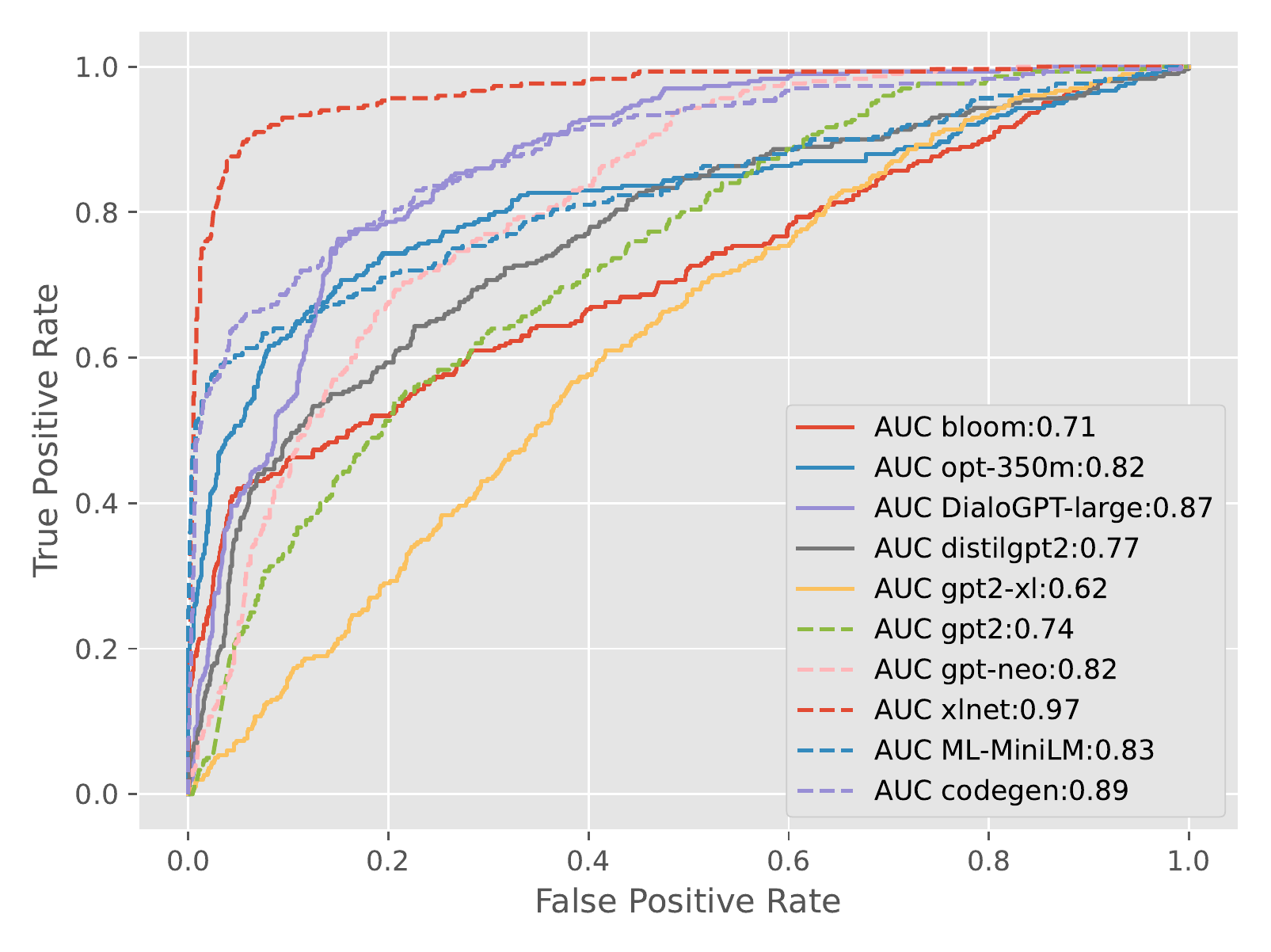}
  \caption{BERT + $I_B$ + P1}
  \label{fig:sub2}
\end{subfigure}%
\begin{subfigure}{.5\textwidth}
  \centering
  \includegraphics[width=0.9\columnwidth]{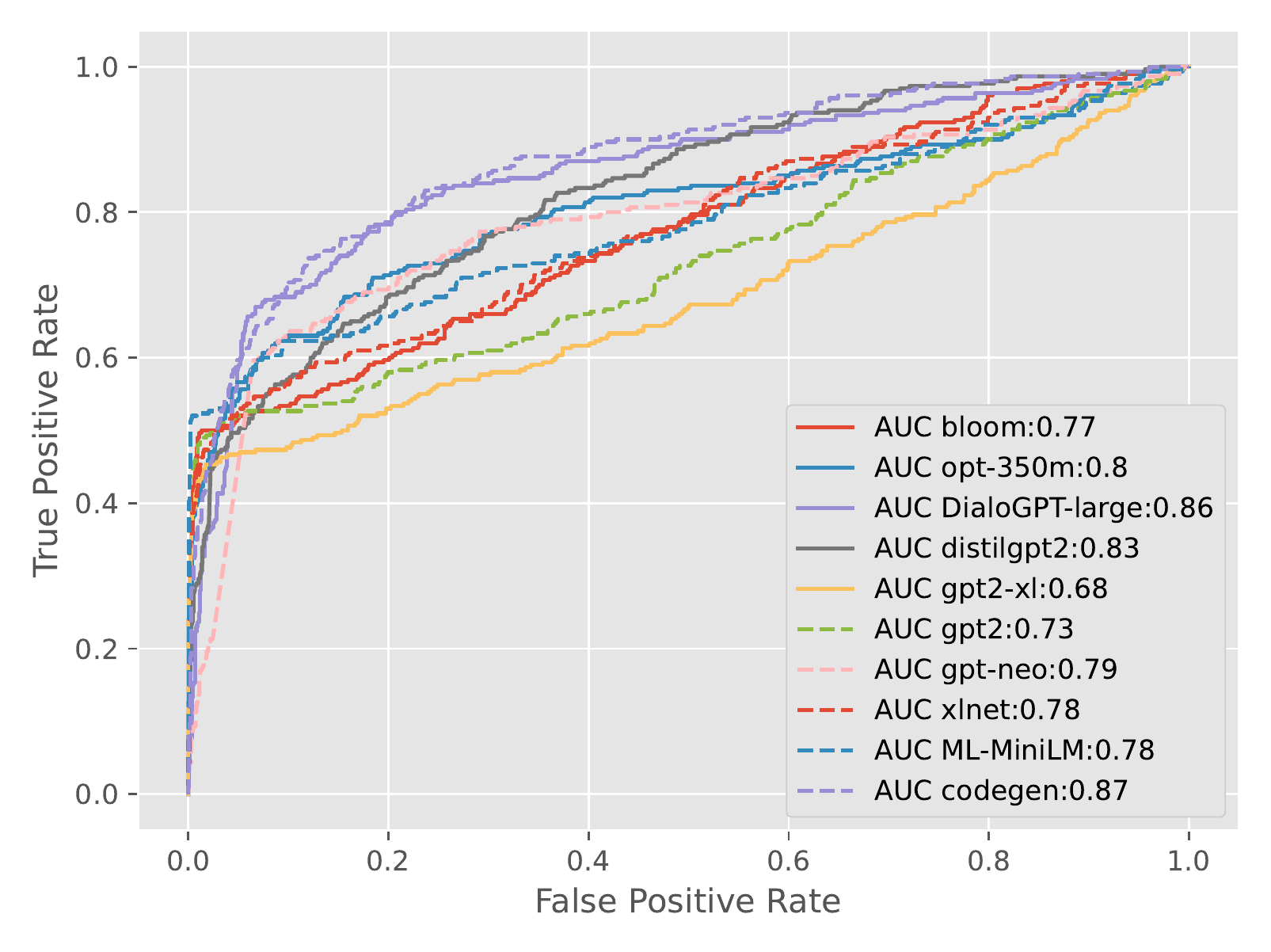}
  \caption{BERT + $I_{B+F}$ + P1}
  \label{fig:sub2}
\end{subfigure}

\begin{subfigure}{.5\textwidth}
  \centering
  \includegraphics[width=0.9\columnwidth]{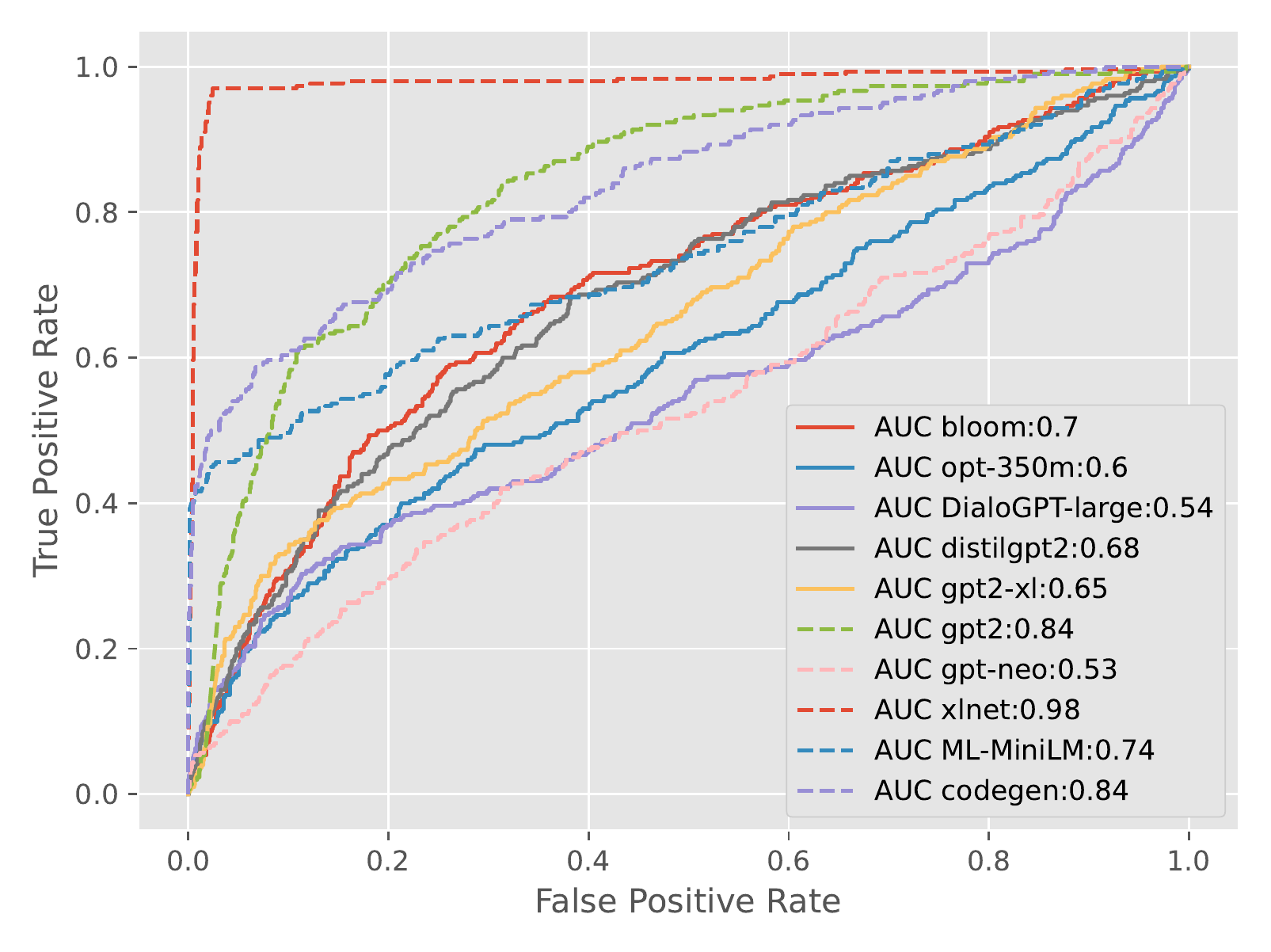}
  \caption{BERT + $I_B$ + P1}
  \label{fig:sub2}
\end{subfigure}%
\begin{subfigure}{.5\textwidth}
  \centering
  \includegraphics[width=0.9\columnwidth]{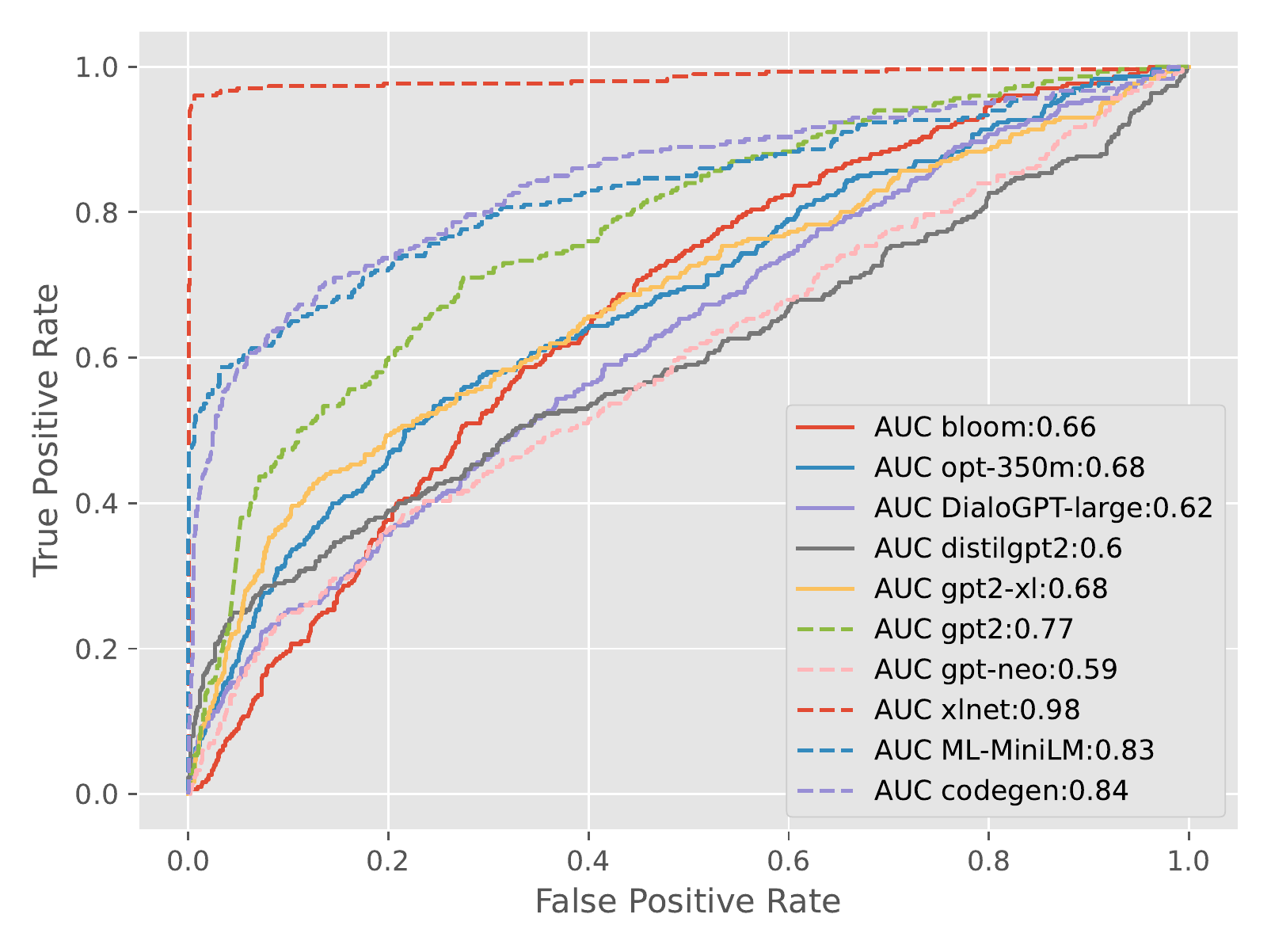}
  \caption{BERT + $I_B$ + P1+P2}
  \label{fig:sub2}
\end{subfigure}
\caption{ROC of Individual base model classifiers, $h_{m_b}$, under different $K$}
\label{fig:roc_auc}
\end{figure*}

\begin{figure*}
\centering
\begin{subfigure}{.5\textwidth}
  \centering
  \includegraphics[width=0.9\columnwidth]{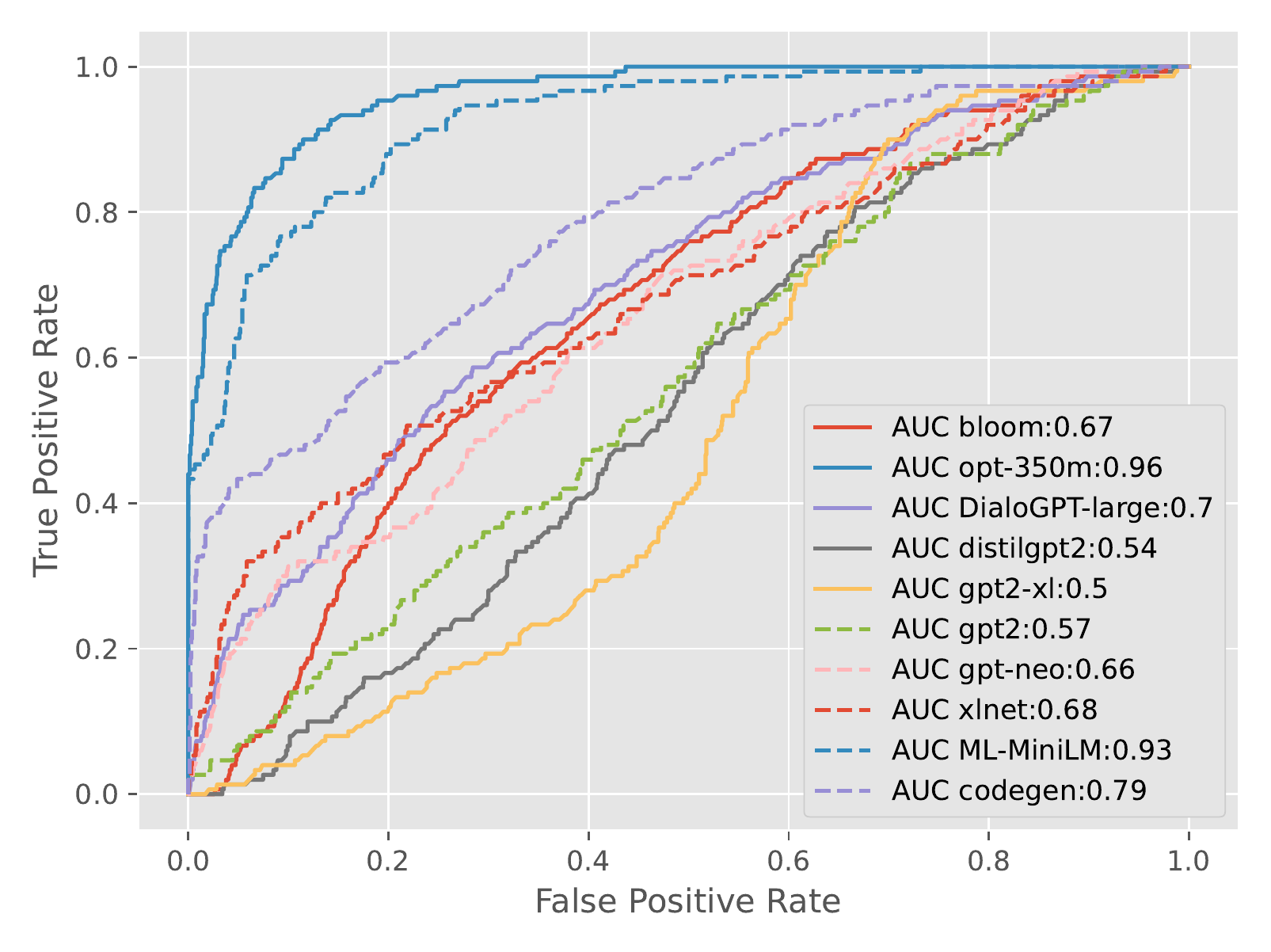}
  \caption{pile150}
  \label{fig:sub1}
\end{subfigure}%
\begin{subfigure}{.5\textwidth}
  \centering
  \includegraphics[width=0.9\columnwidth]{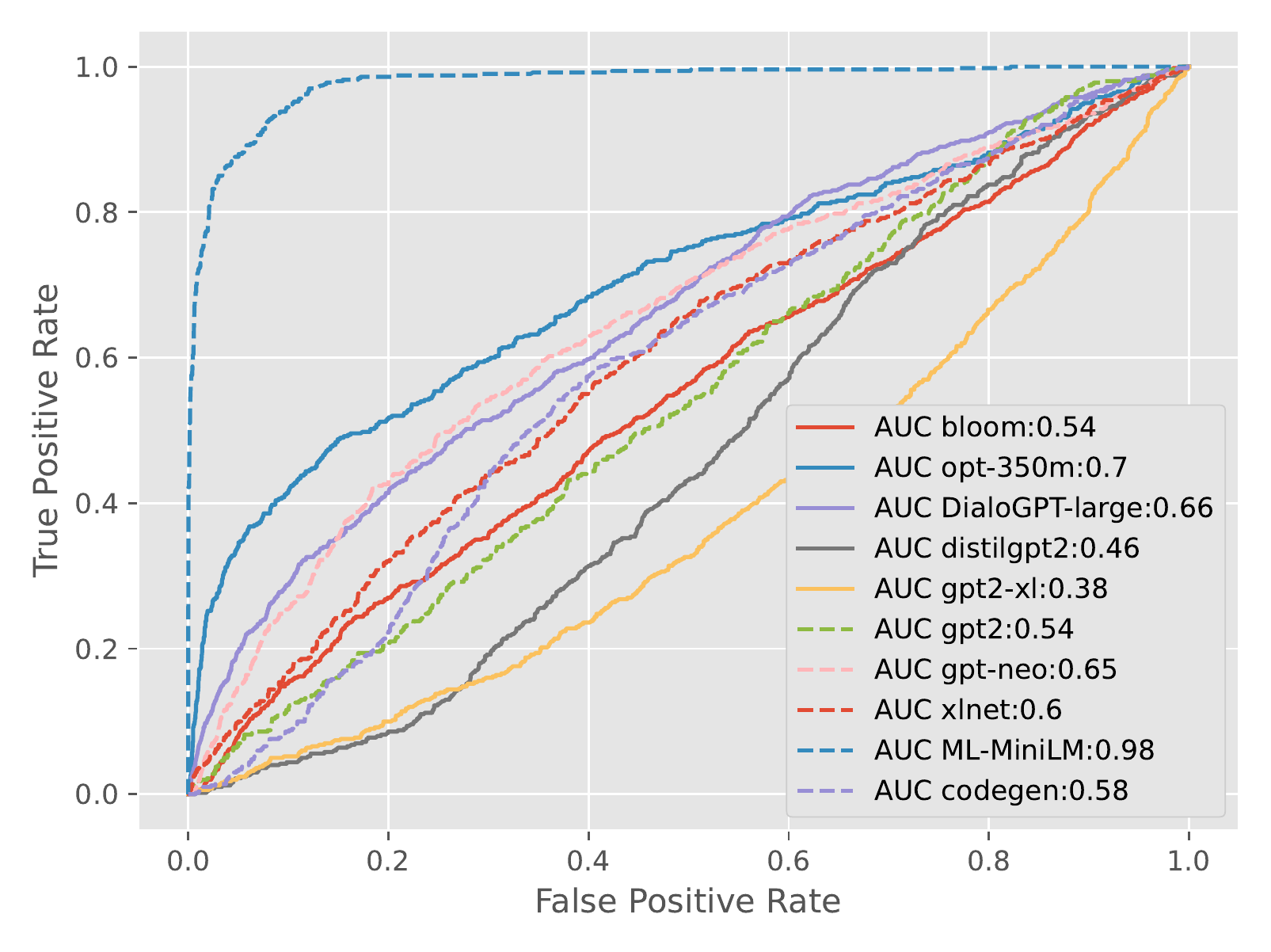}
  \caption{pile 500}
  \label{fig:sub2}
\end{subfigure}

\begin{subfigure}{.5\textwidth}
  \centering
  \includegraphics[width=0.9\columnwidth]{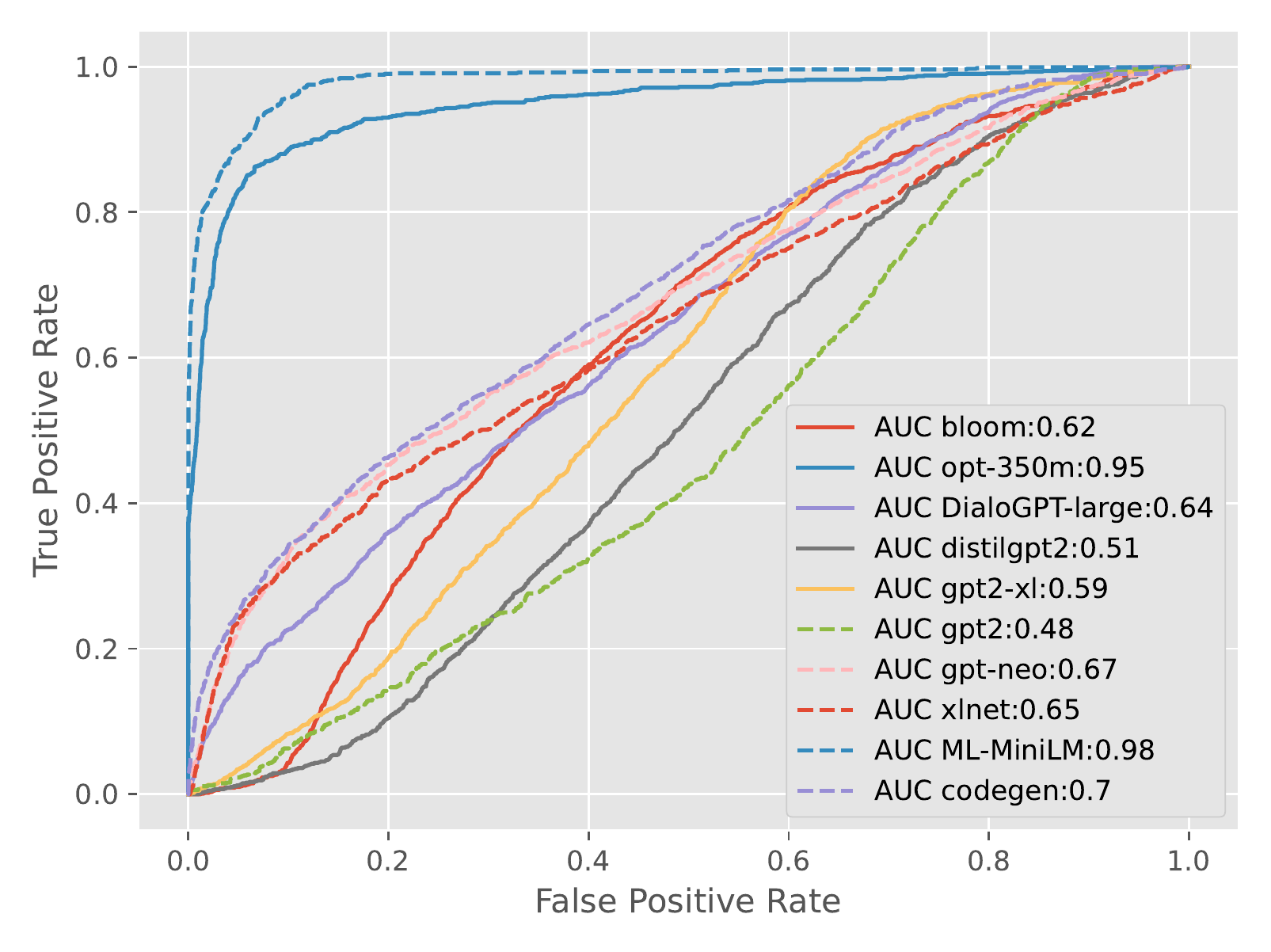}
  \caption{pile 1k}
  \label{fig:sub2}
\end{subfigure}%
\begin{subfigure}{.5\textwidth}
  \centering
  \includegraphics[width=0.9\columnwidth]{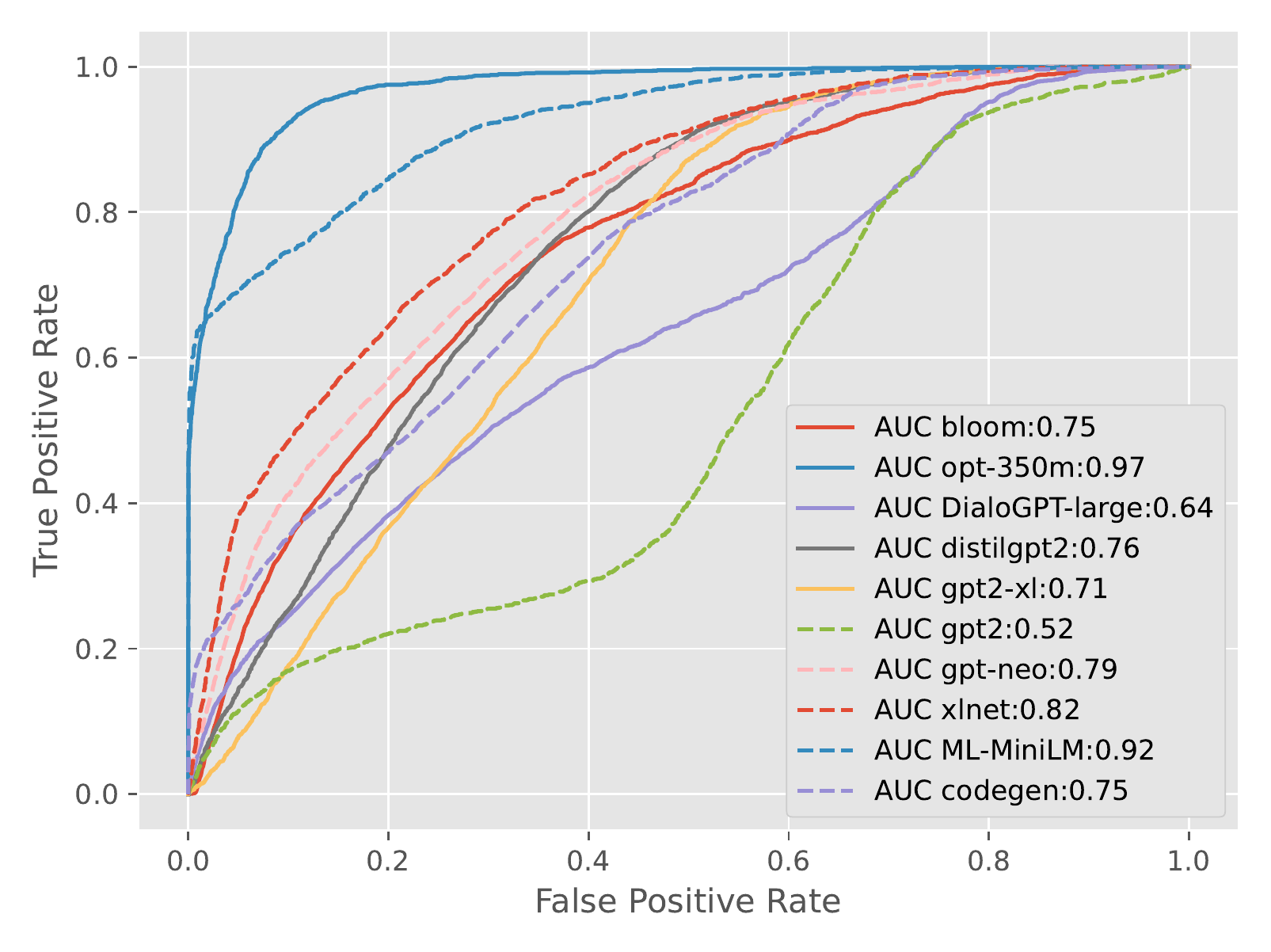}
  \caption{pile 2k}
  \label{fig:sub2}
\end{subfigure}

\begin{subfigure}{.5\textwidth}
  \centering
  \includegraphics[width=0.9\columnwidth]{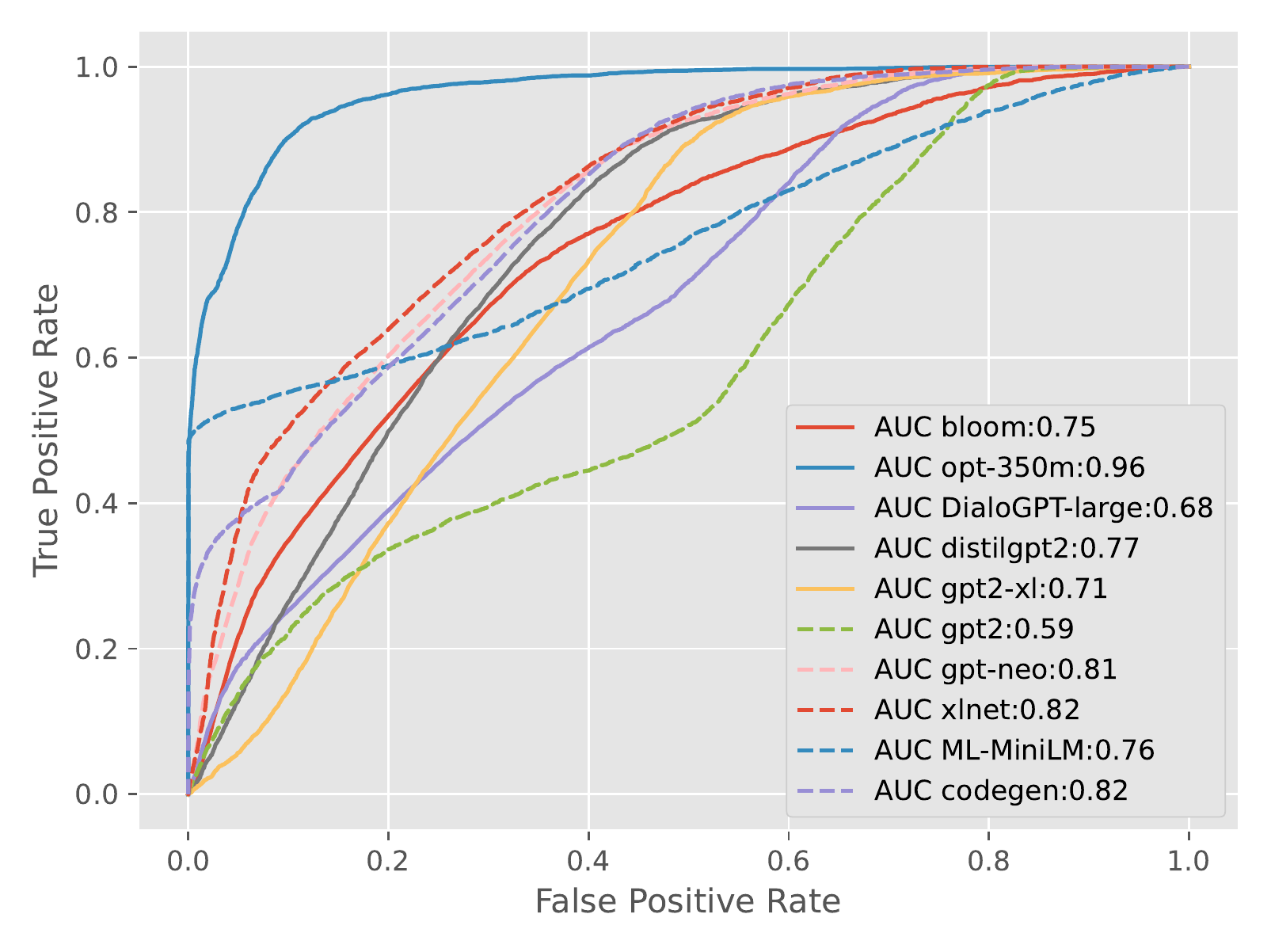}
  \caption{pile 4k}
  \label{fig:sub2}
\end{subfigure}%
\begin{subfigure}{.5\textwidth}
  \centering
  \includegraphics[width=0.9\columnwidth]{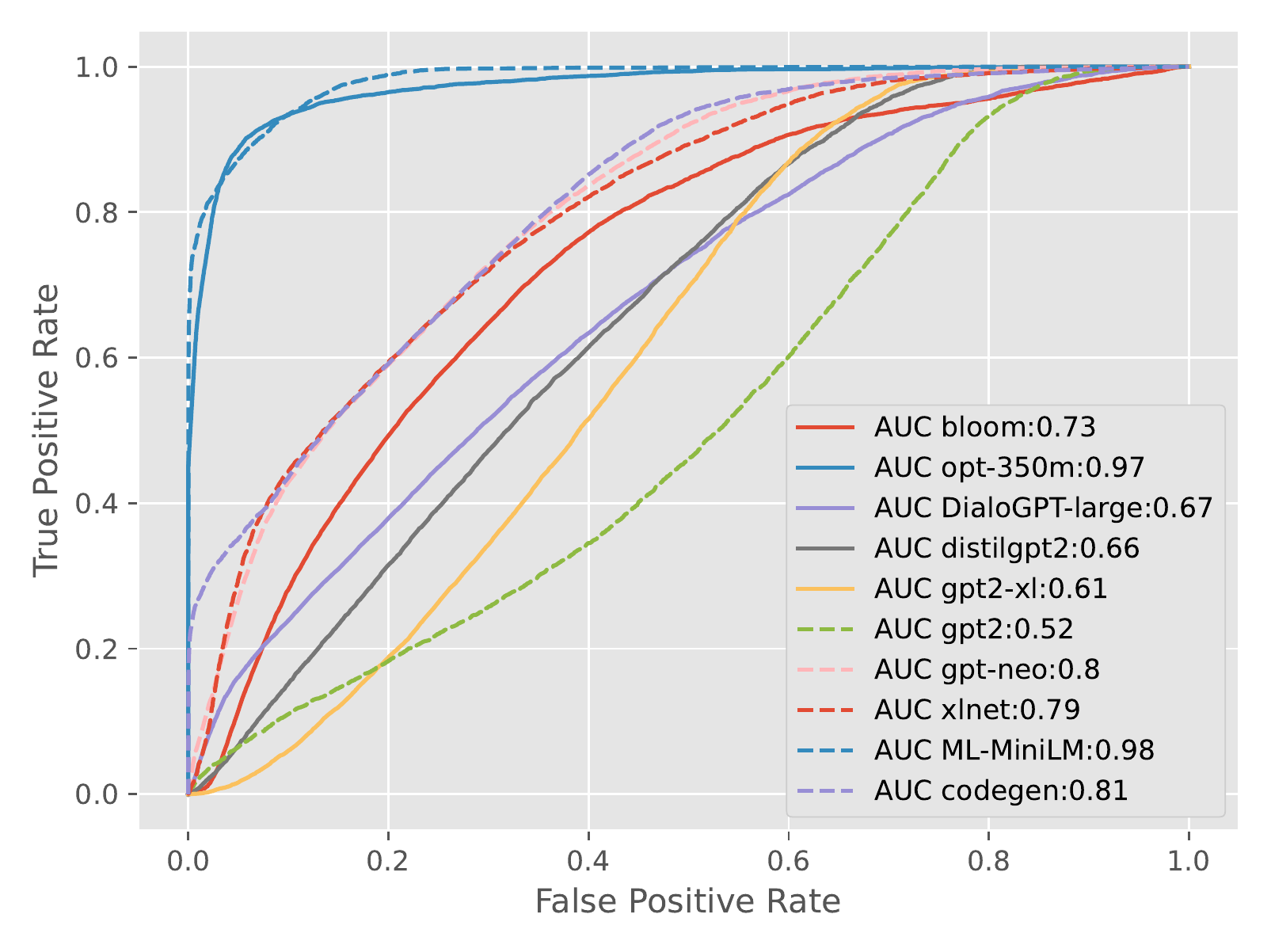}
  \caption{pile 6k}
  \label{fig:sub2}
\end{subfigure}

\begin{subfigure}{.5\textwidth}
  \centering
  \includegraphics[width=0.9\columnwidth]{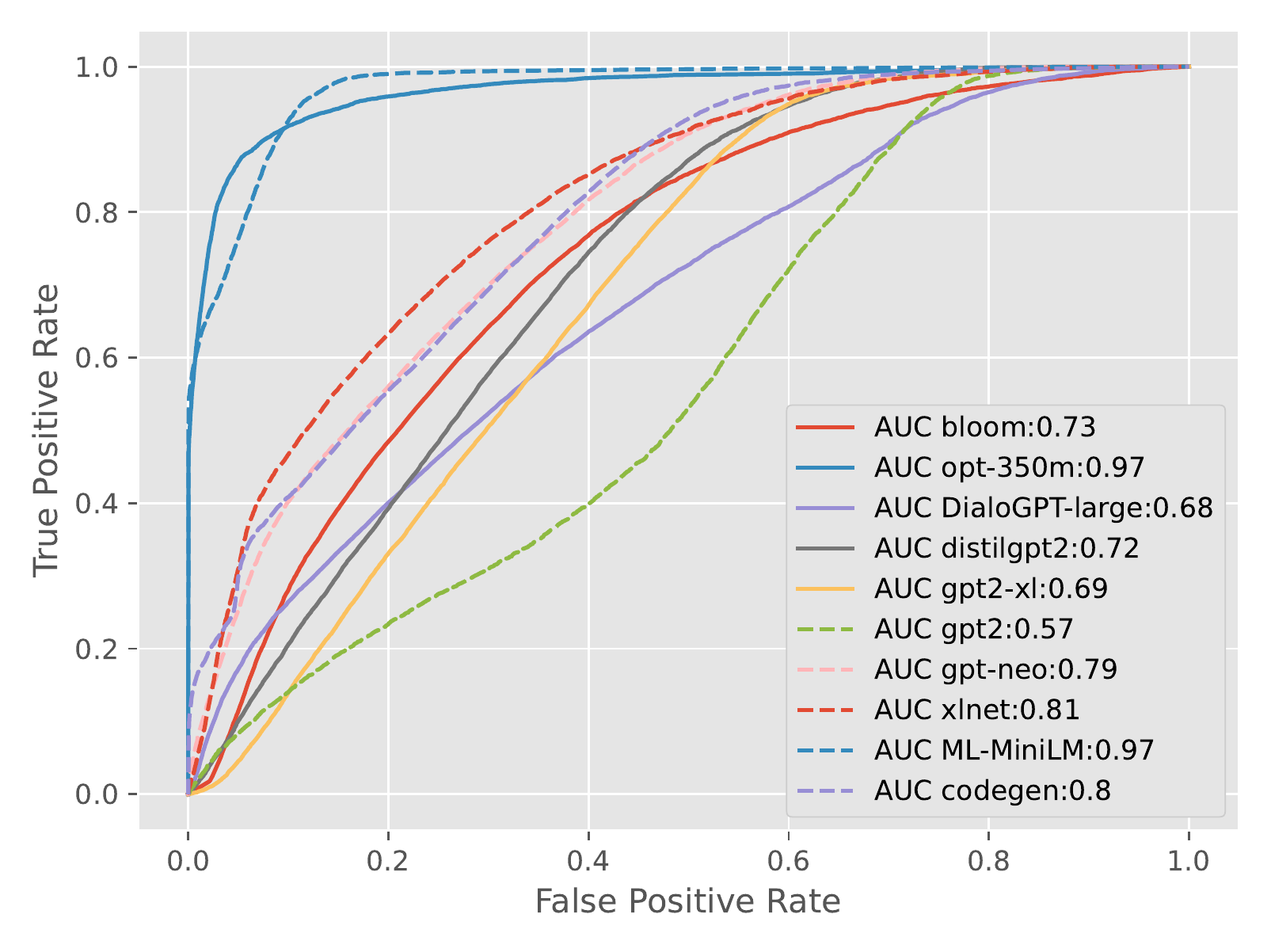}
  \caption{pile 8k}
  \label{fig:sub2}
\end{subfigure}%
\begin{subfigure}{.5\textwidth}
  \centering
  \includegraphics[width=0.9\columnwidth]{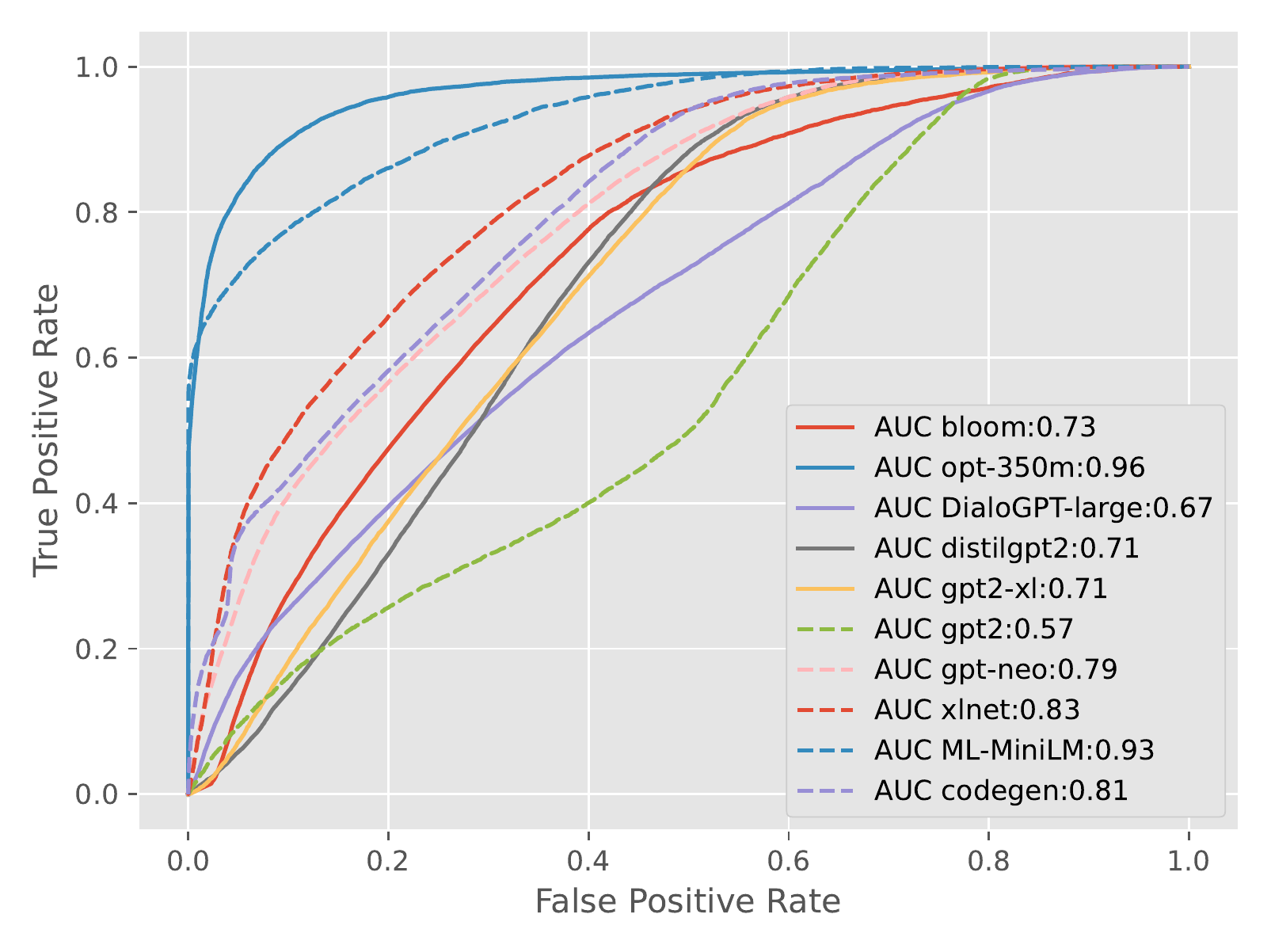}
  \caption{pile 10k}
  \label{fig:sub2}
\end{subfigure}
\caption{ROC of Individual base model classifiers, $h_{m_b}$, with different number of prompts used for attribution under $K_R$}
\label{fig:varying_prompts_all}
\end{figure*}

\begin{figure*}
\centering
\begin{subfigure}{.5\textwidth}
  \centering
  \includegraphics[width=0.9\columnwidth]{figures/auc_roc_all_pile150.pdf}
  \caption{pile150}
  \label{fig:sub1}
\end{subfigure}%
\begin{subfigure}{.5\textwidth}
  \centering
  \includegraphics[width=0.9\columnwidth]{figures/auc_roc_all_pile500.pdf}
  \caption{pile 500}
  \label{fig:sub2}
\end{subfigure}

\begin{subfigure}{.5\textwidth}
  \centering
  \includegraphics[width=0.9\columnwidth]{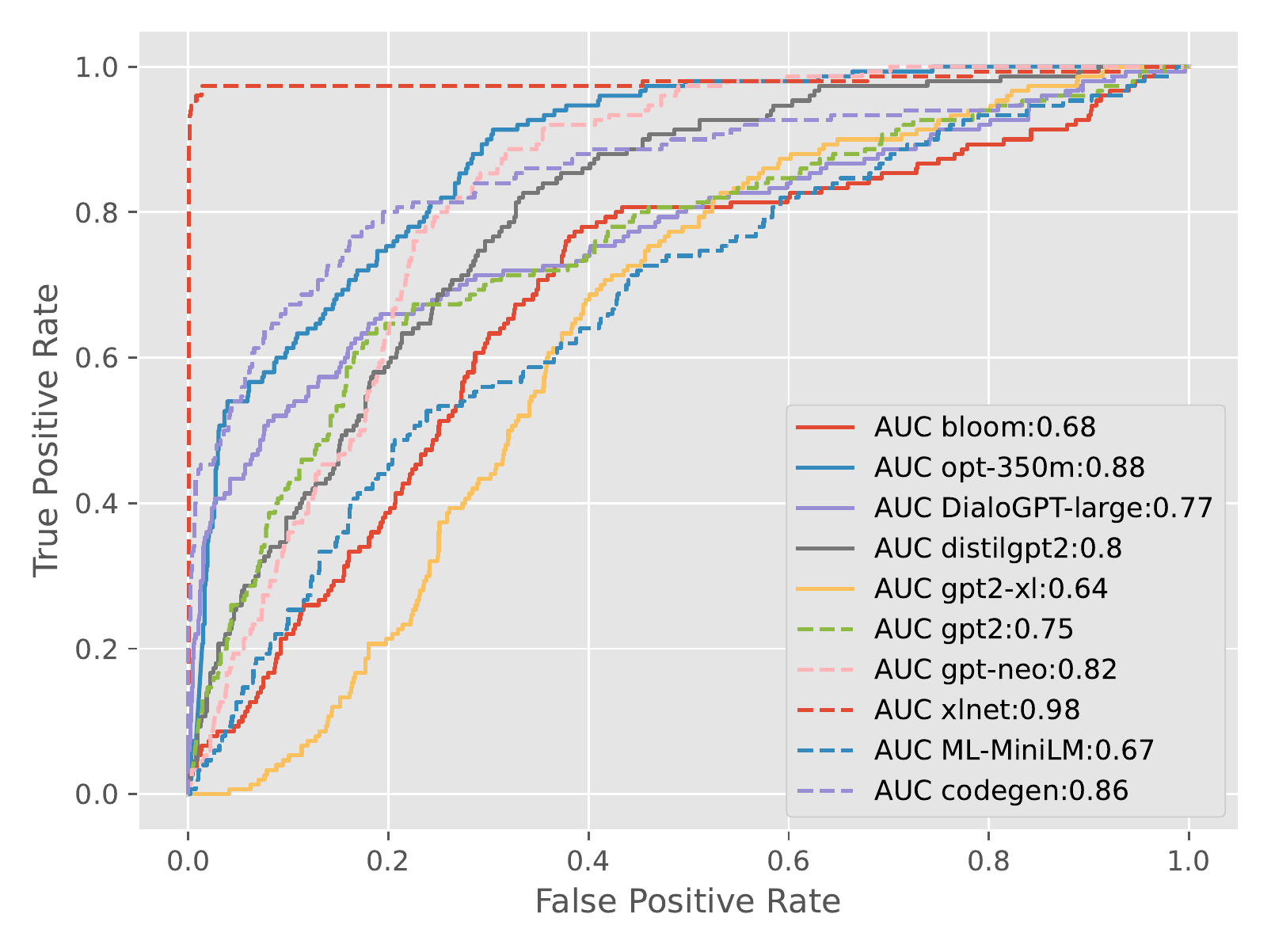}
  \caption{pile 1k}
  \label{fig:sub2}
\end{subfigure}%
\begin{subfigure}{.5\textwidth}
  \centering
  \includegraphics[width=0.9\columnwidth]{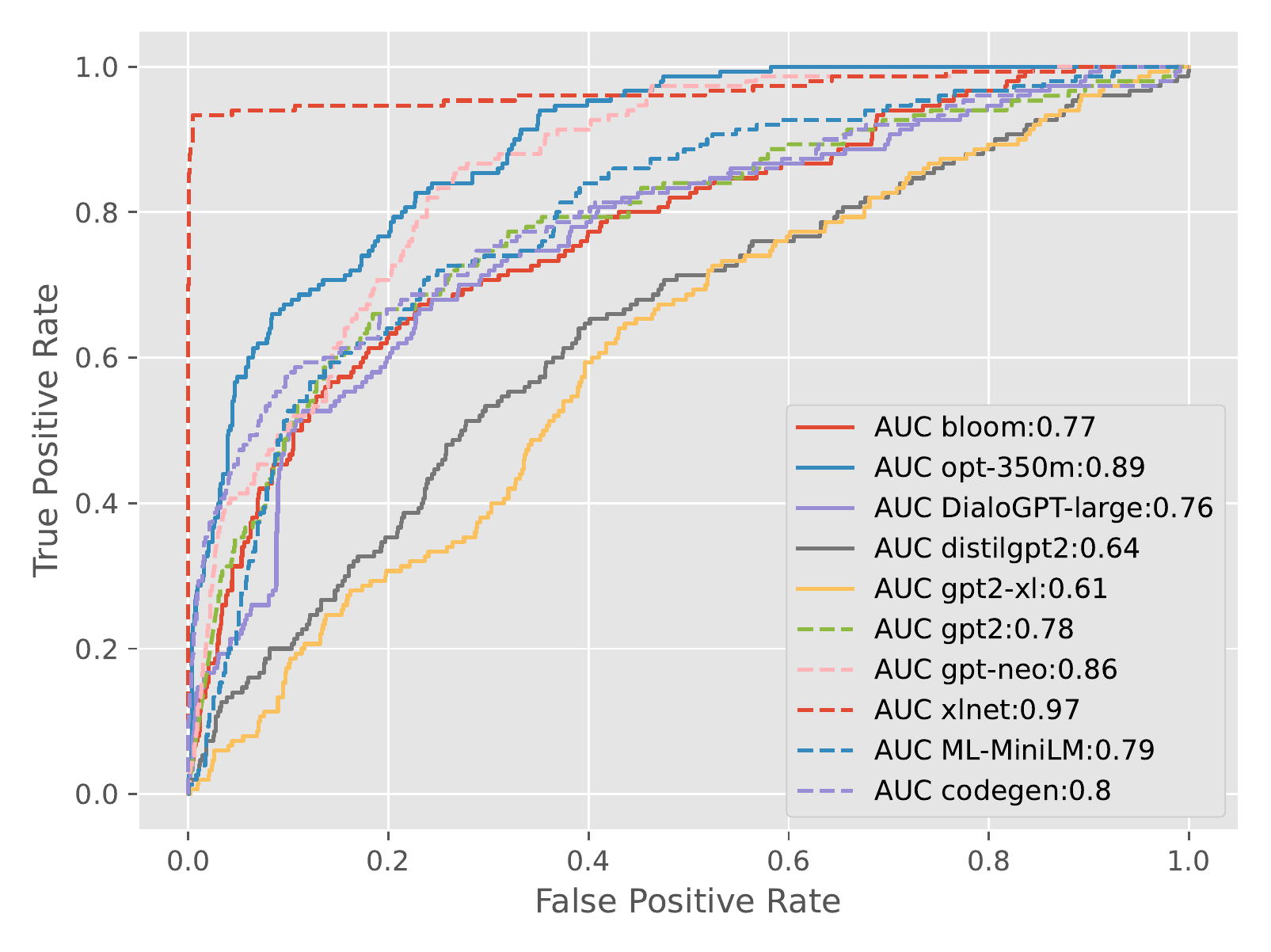}
  \caption{pile 2k}
  \label{fig:sub2}
\end{subfigure}

\begin{subfigure}{.5\textwidth}
  \centering
  \includegraphics[width=0.9\columnwidth]{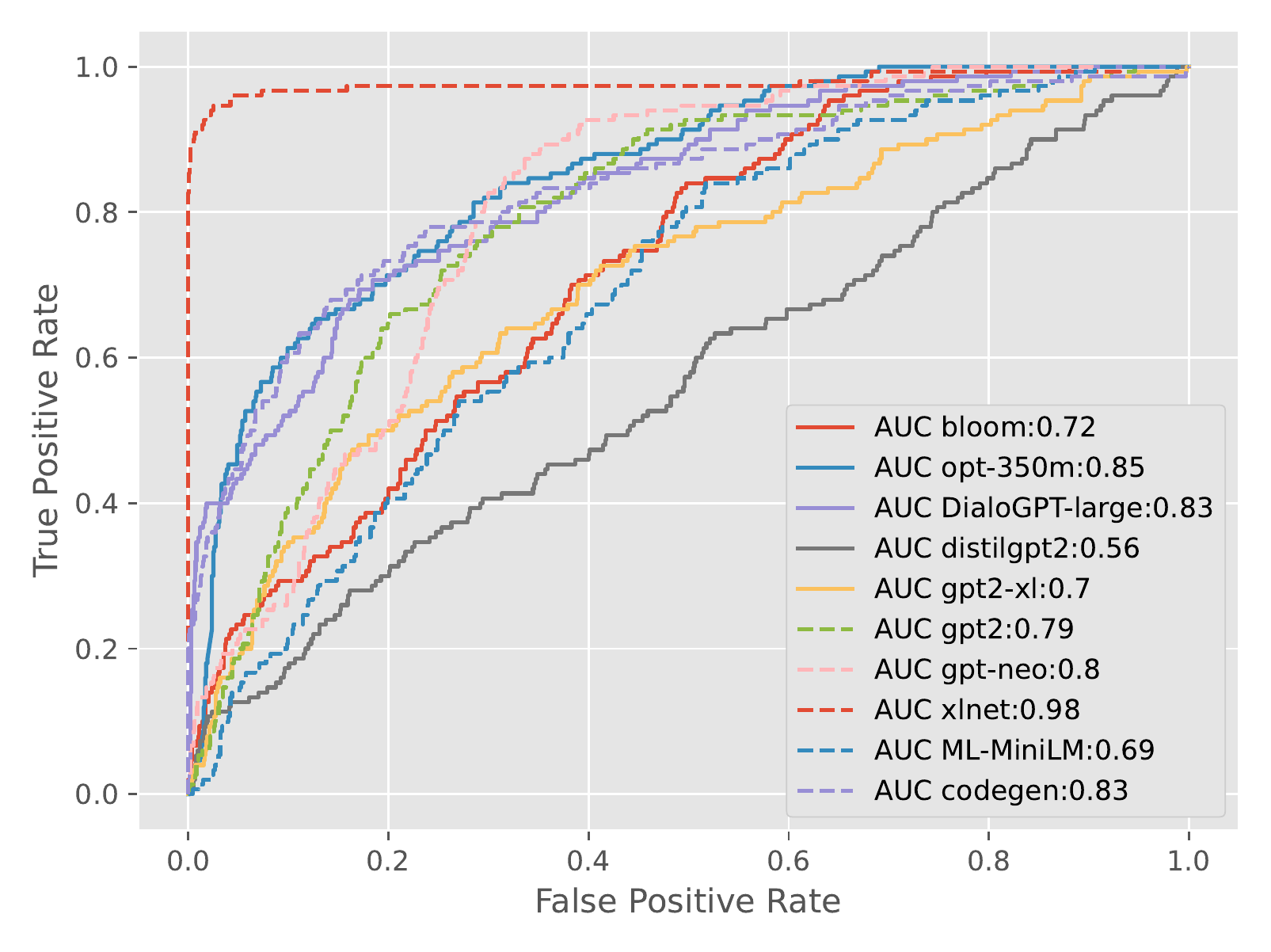}
  \caption{pile 4k}
  \label{fig:sub2}
\end{subfigure}%
\begin{subfigure}{.5\textwidth}
  \centering
  \includegraphics[width=0.9\columnwidth]{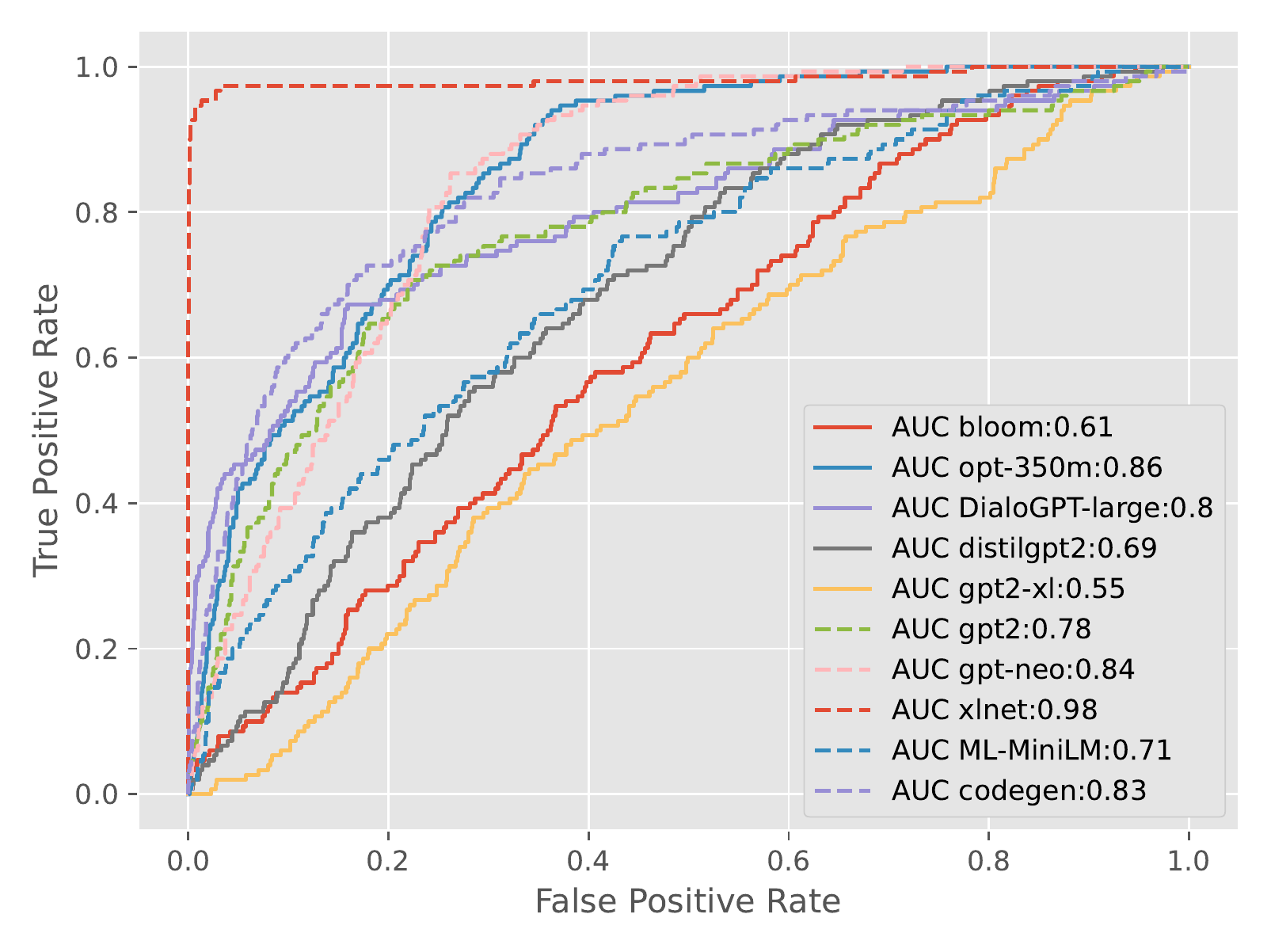}
  \caption{pile 6k}
  \label{fig:sub2}
\end{subfigure}

\begin{subfigure}{.5\textwidth}
  \centering
  \includegraphics[width=0.9\columnwidth]{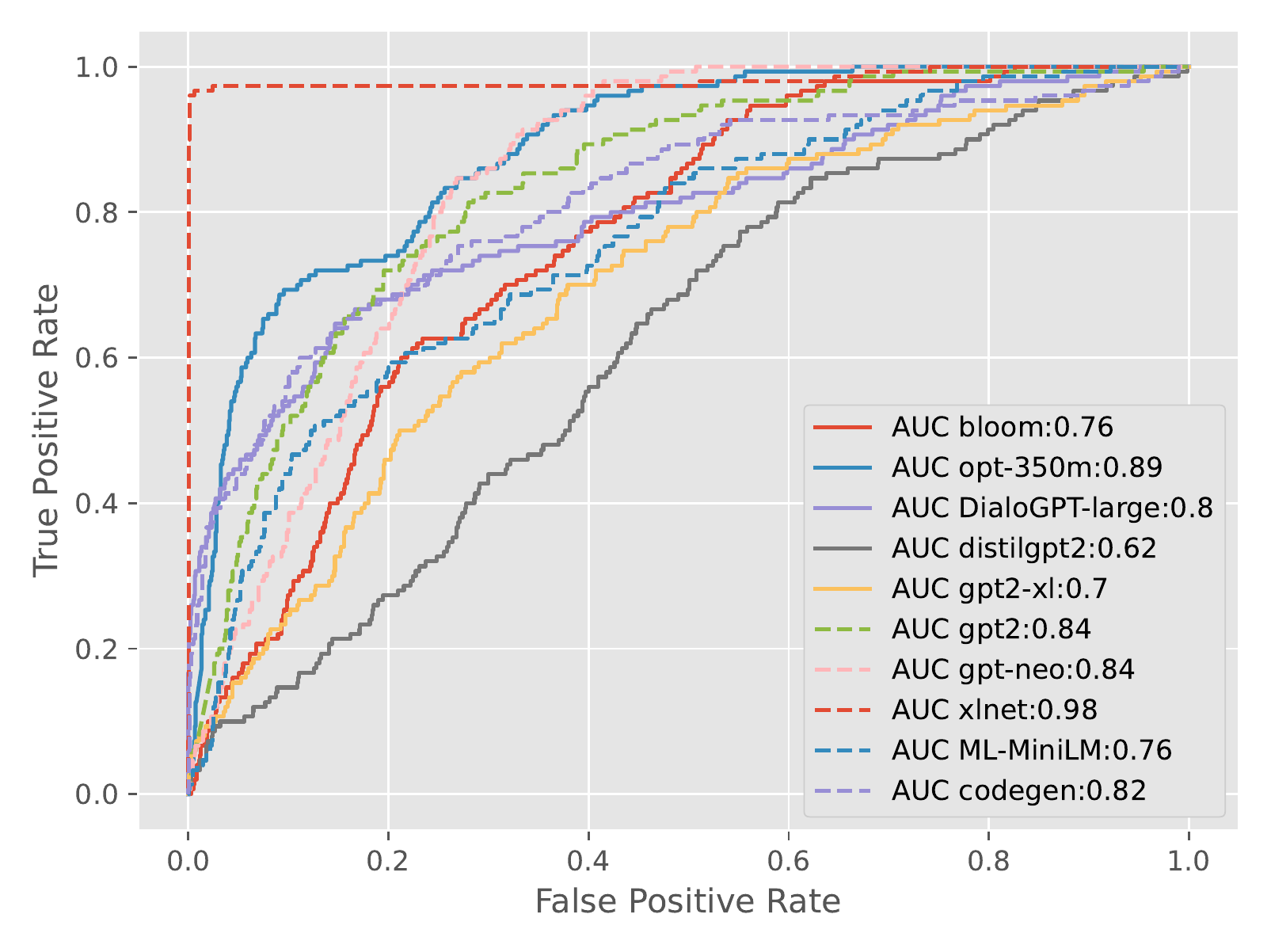}
  \caption{pile 8k}
  \label{fig:sub2}
\end{subfigure}%
\begin{subfigure}{.5\textwidth}
  \centering
  \includegraphics[width=0.9\columnwidth]{figures/auc_roc_all_finetuned_pile10k.pdf}
  \caption{pile 10k}
  \label{fig:sub2}
\end{subfigure}
\caption{ROC of Individual base model classifiers, $h_{m_b}$ using a varying number of prompts for pretraining $h_{m_b}$.}
\label{fig:roc_pretrain_all}
\end{figure*}

We provide the finegrained plots of how each individual $h_{m_b}$ did in each experiment. Figure~\ref{fig:roc_auc} shows the results from the experiment that measures the attribution accuracy under different $K$ as discussed in Section~\ref{ssec:attribtuion_acc}. Figure~\ref{fig:varying_prompts_all} details the effect of using a different number of prompts for attribution under $K_R$, as discussed in Section~\ref{ssec:prompt_pile}. Finally Figure~\ref{fig:roc_pretrain_all} shows the effect of varying the number of prompts for pretaining $h_{m_b}$ (Section~\ref{ssec:pretain_attr}).

\begin{table*}[]\centering {\footnotesize
\begin{tabular}{@{}cc@{}}
\toprule
Dataset & Percentage of prompts in 10,000 subset of the Pile \\ \midrule
Pile-CC & 25.24 \\
OpenWebText2 & 15.20 \\
PubMed Abstracts & 14.23 \\
StackExchange & 13.99 \\
Github & 8.55 \\
Wikipedia (en) & 7.79 \\
USPTO Backgrounds & 5.14 \\
PubMed Central & 2.59 \\
FreeLaw & 2.41 \\
NIH ExPorter & 1.04 \\
DM Mathematics & 0.99 \\
ArXiv & 0.91 \\
HackerNews & 0.81 \\
Enron Emails & 0.47 \\
OpenSubtitles & 0.27 \\
YoutubeSubtitles & 0.11 \\
Books3 & 0.09 \\
EuroParl & 0.06 \\
PhilPapers & 0.05 \\
BookCorpus2 & 0.02 \\
Ubuntu IRC & 0.02 \\
Gutenberg (PG-19) & 0.02
\end{tabular}}\caption{Distribution of the original datasets present in the 10,000 prompt subset of The Pile}\label{tab:pile_dist}
\end{table*}
\section{The Pile subset}\label{app:pile_split}
We make use of a 10,000 prompt subset of The Pile \cite{gao_pile_2020}, in Table~\ref{tab:pile_dist} we report the distrubtion of the smaller datasets present in The Pile.

\end{document}